\def\year{2021}\relax
\documentclass[letterpaper]{article} % DO NOT CHANGE THIS
\usepackage{aaai21}  % DO NOT CHANGE THIS
\usepackage{times}  % DO NOT CHANGE THIS
\usepackage{helvet} % DO NOT CHANGE THIS
\usepackage{courier}  % DO NOT CHANGE THIS
\usepackage[hyphens]{url}  % DO NOT CHANGE THIS
\usepackage{graphicx} % DO NOT CHANGE THIS
\usepackage{natbib}  % DO NOT CHANGE THIS AND DO NOT ADD ANY OPTIONS TO IT
\usepackage{caption} % DO NOT CHANGE THIS AND DO NOT ADD ANY OPTIONS TO IT
\usepackage[utf8]{inputenc} % allow utf-8 input
\usepackage[T1]{fontenc}    % use 8-bit T1 fontsg
\usepackage{booktabs}       % professional-quality tables
\usepackage{amsfonts}       % blackboard math symbols
\usepackage{nicefrac}       % compact symbols for 1/2, etc.
\usepackage{microtype}      % microtypography

\usepackage{tabularx}
\usepackage{graphicx}
\usepackage{amsmath} % for 'cases' env.
\usepackage[utf8]{inputenc}
\usepackage{textcomp}
\usepackage{url}

\usepackage[hang,flushmargin]{footmisc}
\usepackage{color}
\usepackage{xcolor}
\usepackage{makecell}
\usepackage{diagbox}
\usepackage{enumitem}
\usepackage{multirow}
\usepackage{appendix}

\title{Demonstration-Efficient Inverse Reinforcement Learning in Procedurally Generated Environments}
\author{
    Alessandro Sestini,\textsuperscript{\rm 1} 
    Alexander Kuhnle,\textsuperscript{\rm 2} 
    Andrew D. Bagdanov\textsuperscript{\rm 1}
    \\
}
\affiliations{
    \textsuperscript{\rm 1}Dipartimento   di   Ingegneria   dell’Informazione,    Università degli  Studi  di  Firenze, Florence,  Italy\\
    \textsuperscript{\rm 2}Department  of  Computer   Science   and   Technology,   University of   Cambridge, United  Kingdom\\ 
    \{alessandro.sestini, andrew.bagdanov\}@unifi.it, alexander.kuhnle@cantab.net
}

\begin{document}
	\maketitle
	
	\begin{abstract}
        Deep Reinforcement Learning achieves very good results in domains where
        reward functions can be manually engineered. At the same time, there
        is growing interest within the community in using games based on Procedurally Content Generation
        (PCG) as benchmark environments since this type of environment 
        is perfect for
        studying overfitting and generalization of agents under domain shift. Inverse
        Reinforcement Learning (IRL) can instead extrapolate reward functions from
        expert demonstrations, with good results even on high-dimensional problems,
        however there are no examples of applying these techniques to
        procedurally-generated environments. This is mostly due to the number of
        demonstrations needed to find a good reward model. We propose a technique based on
        Adversarial Inverse Reinforcement Learning which can significantly decrease
        the need for expert demonstrations in PCG games. Through the use of an 
        environment with a
        limited set of initial \textit{seed levels}, plus some
        modifications to stabilize training, we show that our approach, DE-AIRL, is demonstration-efficient and still able to extrapolate reward functions which
        generalize to the fully procedural domain. We
        demonstrate the effectiveness of our technique on two procedural
        environments, MiniGrid and DeepCrawl, for a variety of tasks.
	\end{abstract}

\begin{figure*}
	\begin{center}		
	    \includegraphics[width=0.73\textwidth]{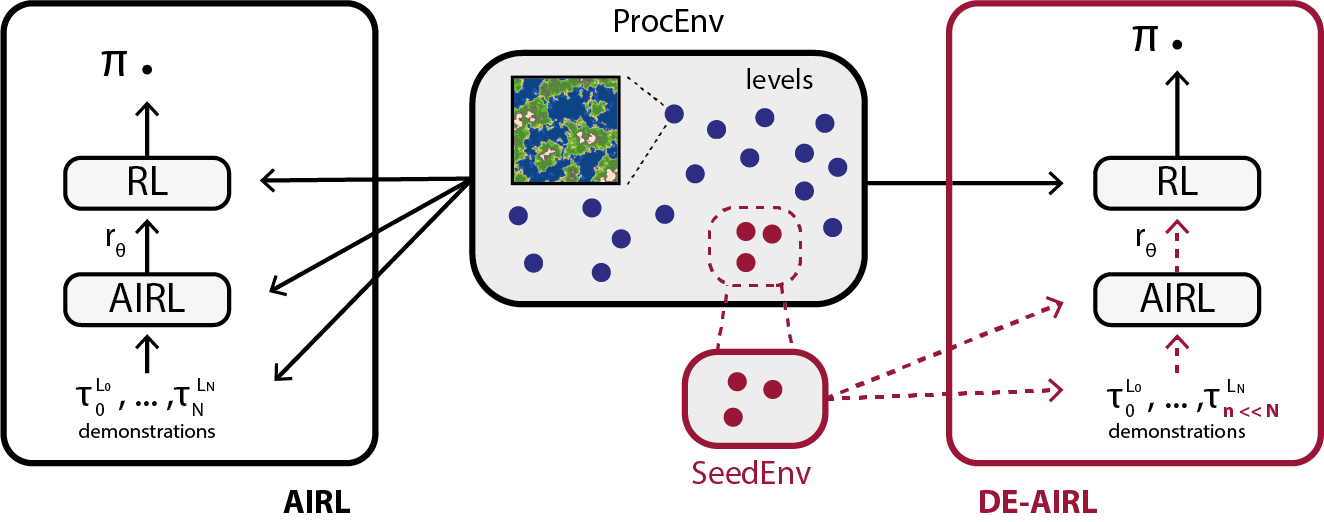} 
	\end{center}
	\caption{Demonstration-efficient AIRL. The left part of the image illustrates the AIRL baseline, which extrapolates a reward
		function from expert demonstrations directly on the fully procedural environment. This naive application of AIRL requires a large number of expert demonstrations. Our demonstration-efficient AIRL approach
		is shown in the right part of the image. DE-AIRL extrapolates the
		reward function on a subset of all possible game levels, referred to as SeedEnv, and is applied in the
		fully procedural environment, ProcEnv, only after training. This approach enables an RL policy to achieve near-expert performance while requiring only a few expert demonstrations.}
	\label{fig:teasing}
\end{figure*}
	
	%%%%%%%%%%%%%%%%%%%%%%%%%%%%%%%%%%%%%%%%%%%%%%%%%%%%%%%%%%%%%%%%%%%%%%%%%%%%%%%%%%%%%%%%%%%%%%%%%%%%%%%%%%%%
	\section{Introduction}
	\label{sec:intro}
	In recent years Deep Reinforcement Learning (DRL) has yielded impressive results
	on problems with known reward functions in complex environments such as video
	games \citep{mnih15, openai2019dota, alphastar} and continuous control
	\citep{levine}. However, designing and engineering good
	hard-coded reward functions is difficult in some domains. In other settings, a
	badly-designed reward function can lead to agents which receive high rewards
	in unintended ways \citep{safety}.
	
	Inverse Reinforcement Learning (IRL) algorithms attempt to infer a reward
	function from expert demonstrations \citep{russelng}. This reward function can
	then be used to train agents which thus learn to mimic the policy implicitly executed by human
	experts. IRL offers the promise of solving many of the problems entailed by
	reward engineering. These approaches have achieved good performance both in
	continuous control tasks \citep{airl,gcl} and in Atari games \citep{adam}.
	
	At the same time, there is increasing interest from the DRL community in
	procedurally-generated environments. In the video game domain, Procedural
	Content Generation (PCG) refers to the programmatic generation of environments
	using random processes that result in an unpredictable and near-infinite range
	of possible states. PCG controls the layout of game levels, the generation of
	entities and objects, and other game-specific details. \citeauthor{procenv} noted that in classical
	benchmarks like the Arcade Learning Environment (ALE) \citep{ale}, agents can
	memorize specific trajectories instead of learning relevant skills, since
	agents perpetually encounter near-identical states. Because of this, PCG
	environments are a promising path towards addressing the need for
	generalization in RL. For an agent to do well in a PCG environment, it has to learn policies robust to ever-changing levels and a general
	representation of the state space.
	
	Most IRL benchmarks focus on finding reward functions in simple and static
	environments like MuJoCo \citep{mujoco} and comparatively simple video games like
	Atari \citep{adam}. None of these RL problems incorporate levels generated
	randomly at the beginning of each new episode. The main challenges with
	procedurally-generated games is the dependence of IRL approaches on the
	number of demonstrations: due to the variability in the distribution of levels,
	if a not sufficiently large number of demonstrations is provided, the reward function
	will overfit to the trajectories in the expert dataset. This leads to an unsuitable
	reward function and consequently poorly performing RL agents. Moreover, in most domains, providing a large
	number of expert demonstrations is expensive in terms of human effort.
	
	To mitigate the need for very many expert demonstrations in PCG games, we
	propose a novel Inverse Reinforcement Learning technique for such
	environments. Our work is based on Adversarial Inverse Reinforcement Learning
	(AIRL) \citep{airl} and substantially reduces the required number of expert trajectories (see
	figure \ref{fig:teasing}). We propose specific changes to AIRL in order to
	decrease overfitting in the discriminator, to increase training stability, and
	to help achieve better performance in agents trained using the learned reward.
	Additionally, instead of using a fully procedural environment for training, we \textit{``under-sample''}
	the full distribution of levels into a small, fixed set of \emph{seed levels}, and experts need only provide demonstrations for this reduced set of
	procedurally-generated levels. We show that the disentangled reward functions learned by AIRL generalize enough such that, subsequently, they enable us to find near-expert policy even on the full distribution of all possible levels. We test
	our approach in two different PCG environments for various tasks.

	%%%%%%%%%%%%%%%%%%%%%%%%%%%%%%%%%%%%%%%%%%%%%%%%%%%%%%%%%%%%%%%%%%%%%%%%%%%%%%%%%%%%%%%%%%%%%%%%%%%%%%%%%%%%
	\section{Related Work}
	\label{sec:related}
	Inverse Reinforcement Learning (IRL) refers to techniques that infer a reward
	function from human demonstrations, which can subsequently be used to train an RL policy. It is often assumed that demonstrations come
	from an expert who is behaving near-optimally. IRL was first described by
	\citeauthor{russelng}, and one of its first successes was by \citeauthor{maxentr} with
	Maximum Entropy IRL, a probabilistic approach based on the principle of maximum
	entropy favoring rewards that lead to a high-entropy stochastic policy. However,
	this approach assumes known transition dynamics and a finite state space, and
	can retrieve only a linear reward function. Guided Cost Learning \citep{gcl}
	relaxed these limitations and was one of the first algorithm able to estimate
	non-linear reward functions over infinite state spaces in environments with
	unknown dynamics. Recently, \citeauthor{gcl} noticed that GCL is closely related
	to GAN training, and this idea led to the development of Adversarial Inverse
	Reinforcement Learning (AIRL) \citep{airl}. This method is able to recover
	reward functions robust to changes in dynamics and can learn policies even under
	significant variations in the environment.
	
	Similarly to IRL, Imitation Learning (IL) aims to directly find a policy that
	mimics the expert behavior from a dataset of demonstrations, instead of
	inferring a reward function which can subsequently be used to train an RL
	policy. Standard approaches are based on Behavioral Cloning \citep{bc1, bc2}
	that mainly use supervised learning \citep{bc1, bc2, dagger, sqil, ilonehot,
		tamer}. Generative Adversarial Imitation Learning (GAIL) \citep{gail} is a
	recent IL approach which is based on a generator-discriminator approach similar
	to AIRL. However, since our goal is to operate in PCG environments, we require
	IRL methods able to learn a reward function which generalizes to different
	levels rather than a policy which tends to overfit to levels seen in expert
	demonstrations.
	
	% \alex{(What's the message here?)} In GAIL, the
	% discriminator outputs the probability that a state-action pair comes from the
	% expert dataset or from the generator. However, the discriminator/critic of GAIL
	% cannot be used as a reward function since, at optimality, it returns 0.5 for
	% all states and actions.
	
	\citeauthor{prefplusimi} combine IL and IRL: they first do an iteration of Behavioral
	Cloning, and then apply active preference learning \citep{pref} in which they
	ask humans to choose the best of two trajectories generated by the policy. With
	these preferences they obtain a reward function, which the policy tries to optimize
	in an iterative process.
	
	Procedural Content Generation (PCG) refers to algorithmic generation of
	level content, such as map layout or entity attributes in video
	games. There is a growing interest in PCG environments from the DRL community.
	As noted above, \citeauthor{procenv} created a suite of PCG benchmarks and
	demonstrated that the ability to generalize becomes an integral component of
	success when agents are faced with ever changing levels. Similarly,
	\citeauthor{procedural} state that often an algorithm will not learn a general policy,
	but instead a policy that only works for a specific version of a specific task
	with specific initial parameters. \citeauthor{illuminating} explored how
	procedurally-generated levels can increase generalization during training,
	showing that for some games procedural level generation enables generalization
	to new levels within the same distribution. Other examples of PCG environments
	used as DRL benchmarks are \citep{nethack, minecraft, minigrid, unitytower,
		sesto19}.
	Notably, \citeauthor{minecraft} applied Imitation Learning in the form of behavioral cloning over a large set of human demonstrations
	in order to improve the sample efficiency of DRL. To the best of our knowledge, our work is the first to apply IRL to
	procedurally-generated environments.
	
	%%%%%%%%%%%%%%%%%%%%%%%%%%%%%%%%%%%%%%%%%%%%%%%%%%%%%%%%%%%%%%%%%%%%%%%%%%%%%%%%%%%%%%%%%%%%%%%%%%%%%%%%%%%%
	\section{Adversarial Inverse Reinforcement Learning}
	\label{sec:background}
	
	Our approach is based on Adversarial Inverse Reinforcement Learning (AIRL),
	which takes inspiration from GANs \citep{gan} by alternating between training a
	discriminator $D_\theta(s,a)$ to distinguish between policy and expert
	trajectories and optimizing the trajectory-generating policy $\pi(a|s)$. The
	AIRL discriminator is given by:
	\begin{equation}
		\label{eq:discriminator}
		D_\theta(s,a) =
		\frac{\exp\{f_{\theta,\omega}(s,a,s')\}}{\exp\{f_{\theta,\omega}(s,a,s')\} + \pi(a | s)},
	\end{equation}
	where $\pi(a|s)$ is the generator policy and
	$f_{\theta,\omega}(s,a,s') = r_\phi(s,a) + \gamma \phi_\omega(s') -
	\phi_\omega(s)$ is a potential base reward function which combines a reward
	function approximator $r(s,a)$ and a reward shaping term $\phi_{\omega}$. For
	deterministic environment dynamics, \citeauthor{airl} show that there is a state-only
	reward approximator $f^*(s,a,s')=r^*(s)+\gamma V^*(s') - V^*(s)=A^*(s,a)$, where
	the reward is invariant to transition dynamics and hence
	\textit{``disentangled''}.
	
	The objective of the discriminator is to minimize the cross-entropy between
	expert demonstrations $\tau^E = (s_0^E, a_0^E, \dots)$ and generated
	trajectories $\tau^\pi = (s_0^\pi, a_0^\pi, \dots)$:
	\begin{align}
		\label{eq:loss_discrim}
		\mathcal{L}(\theta)  = 
		&-E_{\tau^E}\left[ \sum_{t=0}^{T} \log D_\theta(s_t^E,a_t^E) \right] \nonumber \\
		&-E_{\tau^\pi \sim \pi}\left[ \sum_{t=0}^{T} \log \big(1-D_\theta(s_t^\pi,a_t^\pi)\big) \right].
	\end{align}
	
	The authors show that, at optimality, $f^*(s,a) = \log \pi^*(a | s) = A^*(s,a)$,
	which is the advantage function of the optimal policy. The learned reward
	function is based on the discriminator:
	\begin{equation}
		\label{eq:rew_discrim}
		\hat{r}(s,a) = \log(D_\theta(s,a)) - \log(1 - D_\theta(s,a)),
	\end{equation}
	and the generator policy is optimized with respect to a maximum entropy
	objective (using equations (\ref{eq:rew_discrim}) and (\ref{eq:discriminator})):
	\begin{align}
		\label{eq:generator}
		J(\pi)
		&= E_{\tau \sim \pi}\left[ \sum_{t=0}^{T} \hat{r}_t(s_t,a_t) \right] \nonumber  \\
		&= E_{\tau \sim \pi}\left[ \sum_{t=0}^{T} f_\theta(s_t,a_t) - \log(\pi(a_t | s_t)) \right].
	\end{align}
	
	%%%%%%%%%%%%%%%%%%%%%%%%%%%%%%%%%%%%%%%%%%%%%%%%%%%%%%%%%%%%%%%%%%%%%%%%%%%%%%%%%%%%%%%%%%%%%%%%%%%%%%%%%%%%
	\section{Modifications to AIRL}
	\label{sec:modifications}
	
	In the following we present three extensions to the original AIRL algorithm
	which increase stability and performance, while decreasing the tendency of the
	discriminator to overfit to the expert demonstrations.
	\begin{itemize}

    	\item{\textbf{Reward standardization.} Adversarial training alternates between
    	discriminator training and policy optimization, and the latter is conditioned on
    	the reward which is updated with the discriminator. However, forward RL assumes
    	a stationary reward function, which is not true in adversarial IRL training.
    	Moreover, policy-based DRL algorithms usually learn a value function based on
    	rewards from previous iterations,
    	which consequently may have a different scale from the currently observed
    	rewards due to discriminator updates. Generally, forward RL is very sensitive to
    	reward scale which can affect the stability of training. For these reasons, as
    	suggested in \citeauthor{adam} and \citeauthor{prefplusimi}, we standardize the reward to
    	have zero mean and some standard deviation.}
    	
    	\item{\textbf{Policy dataset expansion.} In the original AIRL algorithm, each
    	discriminator training step is followed by only one policy optimization step.
    	The experience collected in this policy step is then used for the subsequent
    	discriminator update. However, a single trajectory may not offer enough data
    	diversity to prevent the discriminator from overfitting. Hence, instead of just
    	one policy step, we perform \textit{K} iterations
    	of forward RL for every discriminator step as suggested by \citeauthor{adam}.}
    	%\alex{(removed: based on the guided cost learning algorithm \citep{gcl})}
    	
    	Moreover, as already noted by \citeauthor{airl} and \citeauthor{sqil}, IRL methods tend to
    	learn rewards which explain behavior locally for the current policy, because the
    	reward can forget the signal that it gave to an earlier policy. To mitigate this
    	effect we follow their practice of using experience replay over the previous
    	iterations as policy experience dataset. For the same reason, when we
    	apply the learned reward function, we do not re-use the final, possibly
    	overfitted reward model, but rather one from an earlier training iteration.
    	
    	\item{\textbf{Fixed number of timesteps.} Many environments have a terminal
    	condition which can be triggered by agent behavior. \citeauthor{pref} observed that
    	these conditions can encode information about the environment even when the
    	reward function is not observable, thus making the policy task easier.
    	Moreover, since the range of the learned reward model is arbitrary, rewards may
    	be mostly negative in some situations, which encourages the agent to meet the
    	terminal conditions as early as possible to avoid more negative rewards (the
    	so-called \textit{``RL suicide bug''}). For these reasons we do not terminate an
    	episode in a terminal state, but artificially extend it to a fixed number of
    	timesteps by repeating the last timestep.}
	
	\end{itemize}
	
	\begin{figure*}
		\begin{center}
			\scalebox{1.0}{
				\begin{tabular}{cccc}
					\includegraphics[width=0.21\textwidth]{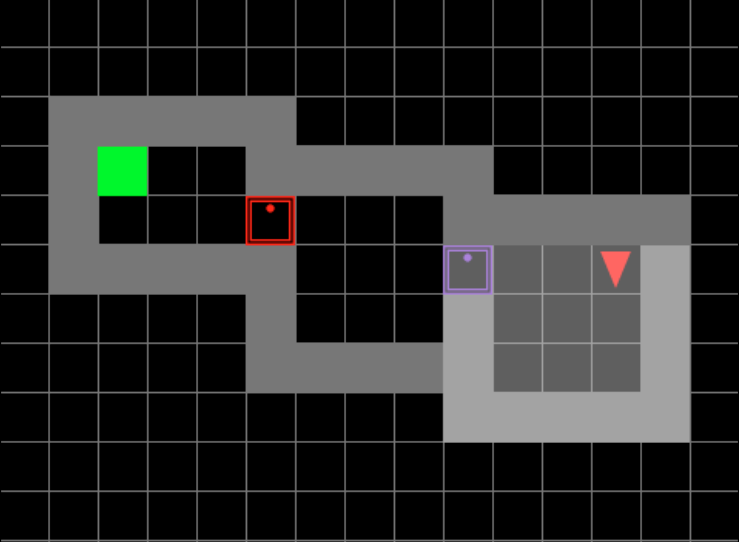} &
					\includegraphics[width=0.21\textwidth]{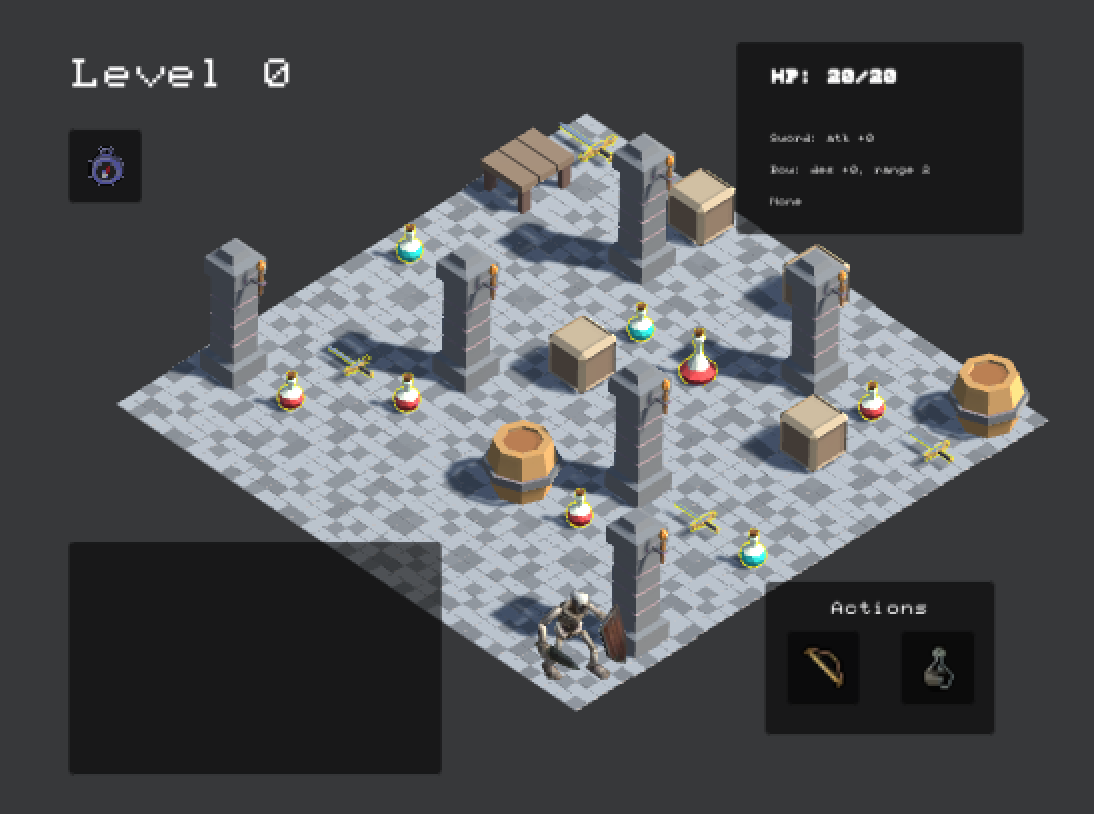} &
					\includegraphics[width=0.21\textwidth]{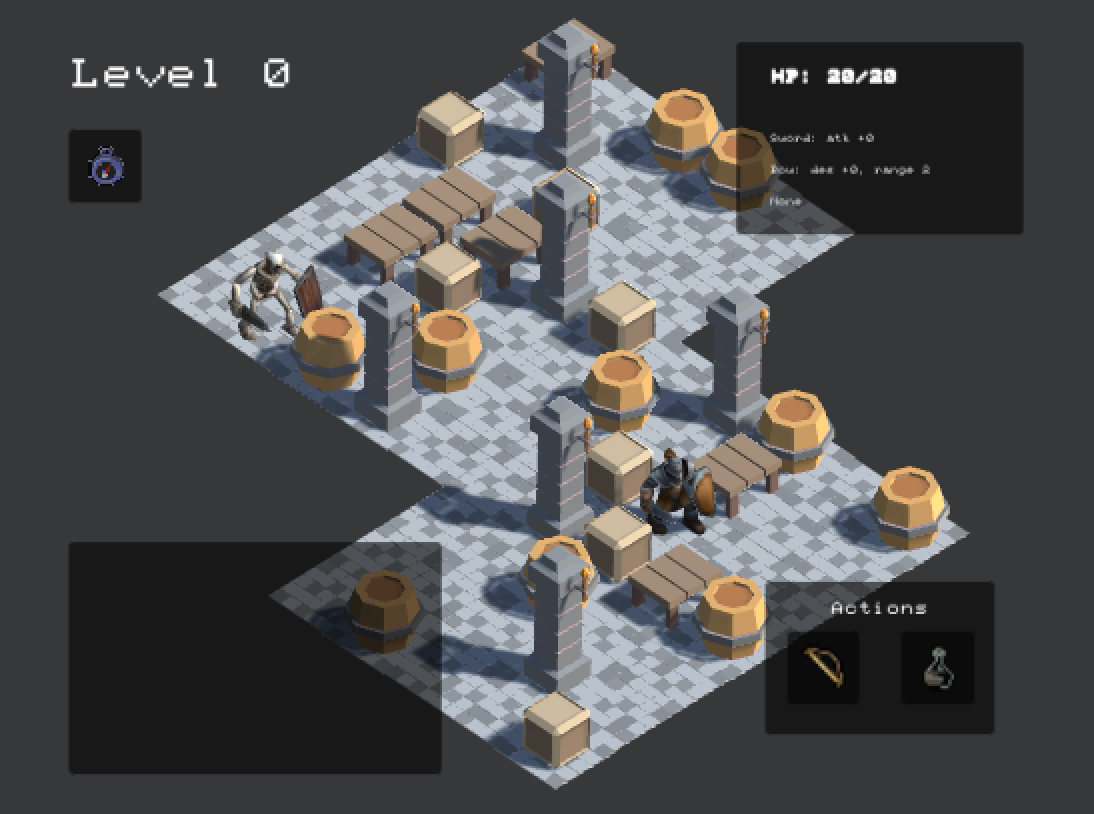} &
					\includegraphics[width=0.21\textwidth]{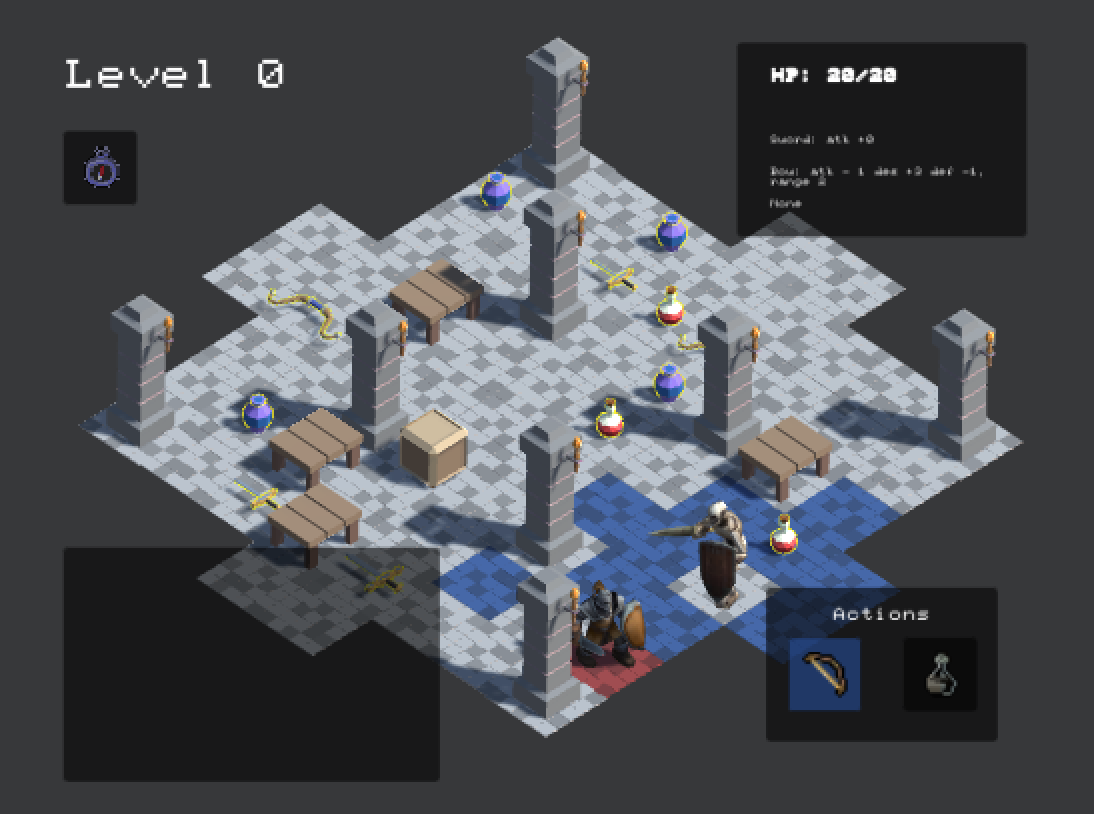} \\
					
					(a) Minigrid & 
					(b) Potions task &
					(c) Maze task &
					(d) Ranged task \\
					
			\end{tabular}}
			
		\end{center}
		\caption{
			Screenshots of the various environments and tasks.
		}
		\label{fig:screen}
	\end{figure*}

	%%%%%%%%%%%%%%%%%%%%%%%%%%%%%%%%%%%%%%%%%%%%%%%%%%%%%%%%%%%%%%%%%%%%%%%%%%%%%%%%%%%%%%%%%%%%%%%%%%%%%%%%%%%%
	\section{Demonstration-efficient AIRL in procedural environments}
	\label{sec:seedmethod}
	
	In PCG game environments, the configuration of the level as well as its entities
	are determined algorithmically. Unless the game is very simplistic, this means
	it is unlikely to see the exact same level configuration twice. Forward RL
	benefits from such environmental diversity by increasing the level of
	generalization and credibility of agent behavior. However, as a consequence of
	this diversity, many expert demonstrations may be required for IRL to learn
	useful behavior. This is especially challenging for an adversarial techniques
	like AIRL as it is known that GANs require many of positive examples
	\citep{gandataset}.
	
	\begin{figure*}
    	\begin{center}
    		\scalebox{0.85}{ \setlength\tabcolsep{-6pt}
    			\begin{tabular}{ccc}
    				\includegraphics[width=0.37\textwidth]{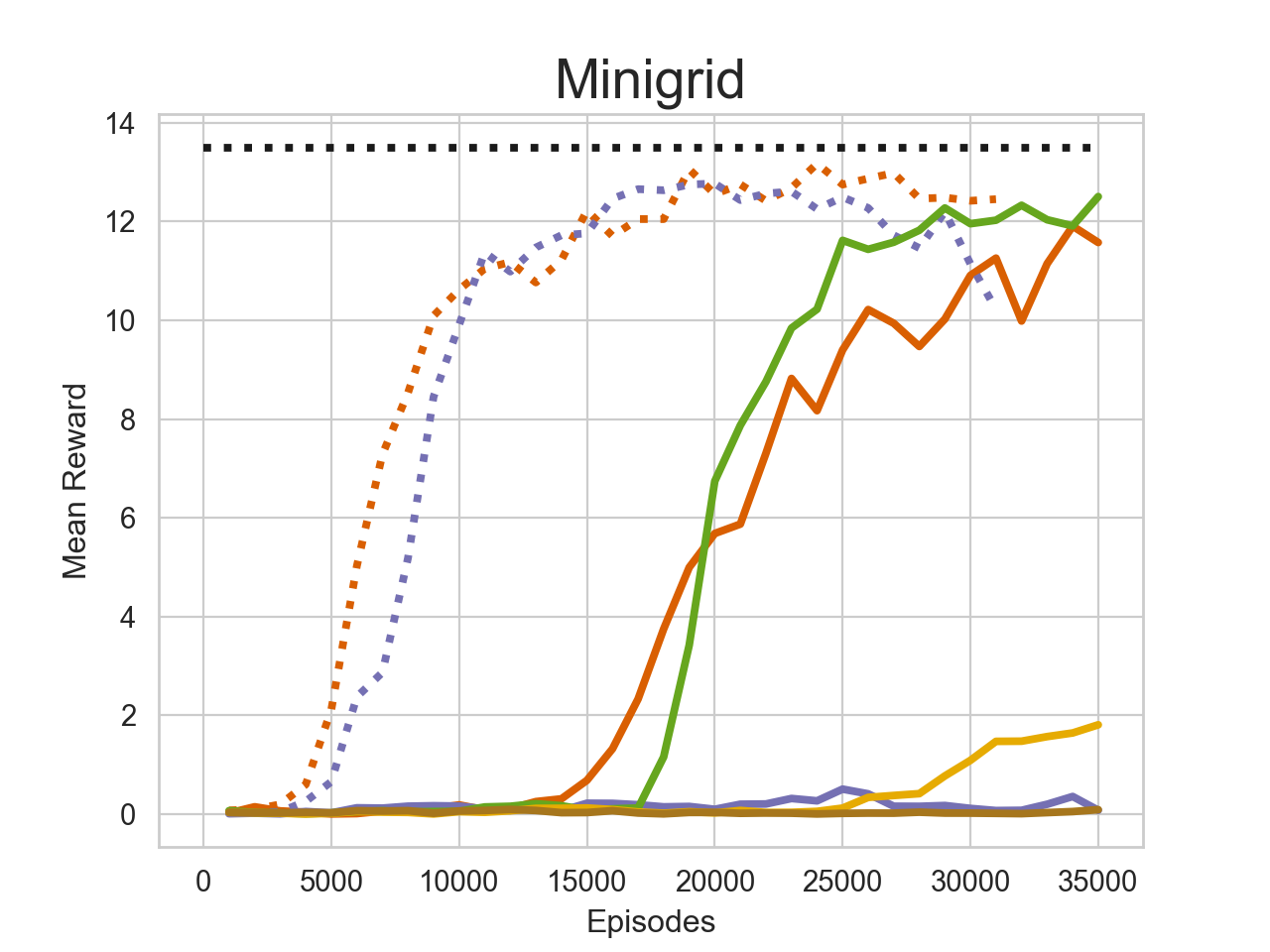} &
    				\includegraphics[width=0.37\textwidth]{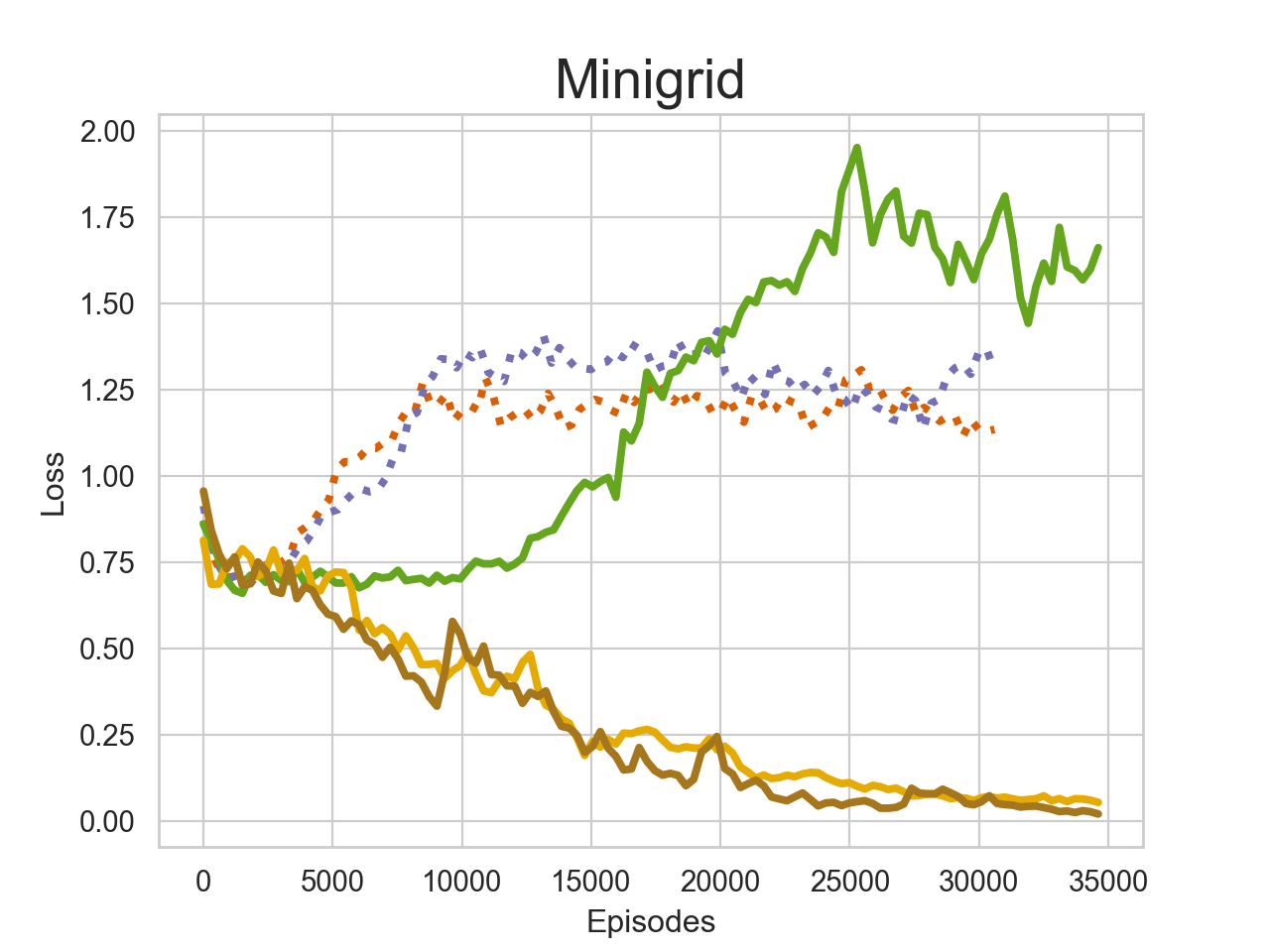} &
    				\includegraphics[width=3.5cm]{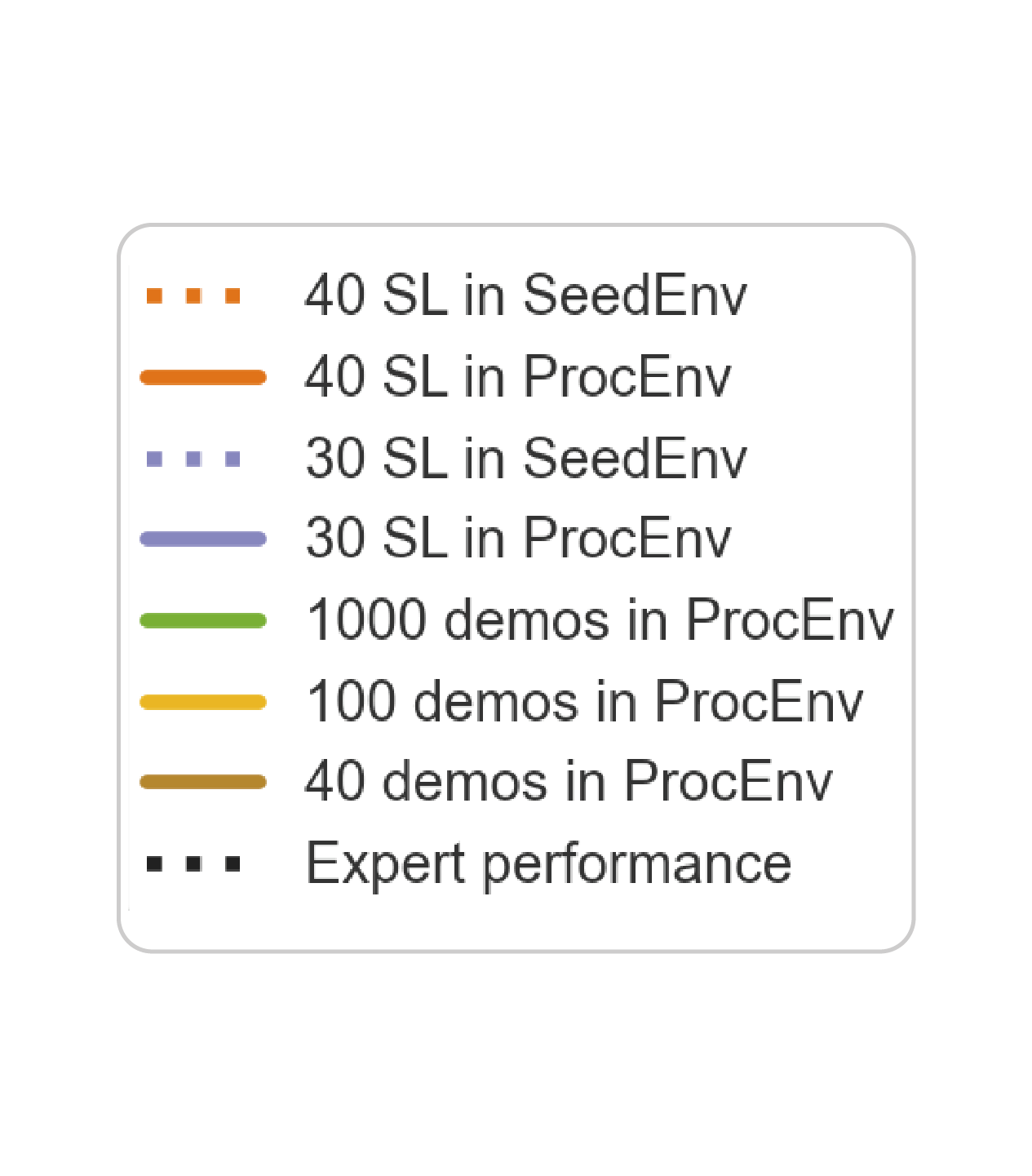} 
    				\\
    				
    				(a) & (b) & \\
    				
    			\end{tabular}
    		}
    	\end{center}
    	\caption{Experimental results on MiniGrid. (a) Mean reward during training
    		for: our DE-AIRL with different numbers of seed levels on both SeedEnv and
    		ProcEnv, naive AIRL with different numbers of demonstrations on ProcEnv, and
    		expert performance on ProcEnv. (b) Discriminator loss during training on
    		either SeedEnv (our approach) or ProcEnv (naive AIRL).}
    	\label{fig:minigrid_results}
    \end{figure*}
	
	In the following, we call the fully procedural environment
	\textbf{ProcEnv}. Levels $L_i \sim \mbox{ProcEnv}$ are sampled from
	this environment, and sample trajectories $\tau^{L_i} \sim L_i$ from each level,
	where trajectories $\tau^{L_i} = (s_0,a_0,\dots,a_{T-1},s_T)$ are sequences of
	alternating states and actions. Consequently, if we have two trajectories
	$\tau^{L_i}$ and $\tau^{L_j}$, in most cases (unless $L_i = L_j$) they differ not only
	in their state-action sequences, i.e. the behavior, but also in their level
	content $L_i$ vs $L_j$ from which they are sampled.
	
	To illustrate this, suppose we have a simple ProcEnv with two
	generation parameters: the number of objects $o \in [1,10]$ and the number of
	enemies $e \in [1,6]$, so overall $|\mbox{ProcEnv}| = 10 \cdot 6 = 60$ level
	configurations. Sampling expert demonstrations is a two-stage process: first, we
	sample levels $L_1=(3,4), L_2=(5,1), L_3=(7,2) \sim \mbox{ProcEnv}$, where
	$(o,e)$ denotes the number of objects and enemies, respectively, and next we
	sample corresponding trajectories $\tau_1^{(3,4)}, \tau_2^{(5,1)}, \tau_3^{(7,2)}$, which
	form our expert dataset. When faced with another trajectory sample based on a
	random level, say, $\tau^{(1,4)}$, the discriminator can simply distinguish expert
	and non-expert trajectories by counting objects and enemies in the levels as
	observed in the states of the trajectories and \emph{ignoring} agent behavior
	entirely. Sampling more expert trajectories increases the probability of levels
	being equal (or at least similar), and thus makes it harder to memorize level
	configurations. However, collecting a large number of demonstrations can be very
	expensive, and cannot not ultimately solve the problem for rich enough PCG
	environments.

	Our objective is to make AIRL effective and data-efficient when working with PCG
	environments. Our main idea is to introduce an artificially reduced environment,
	which we call a \textbf{SeedEnv}, that consists of $n \ll N$ levels
	sampled from the fully procedural ProcEnv. These levels are then used
	to obtain $n$ randomly sampled expert demonstrations:
	\begin{gather*}
		\mbox{SeedEnv}(n) = \{L_1, ..., L_n \mid L_i \sim \mbox{ProcEnv}\}  \\
		\text{Demos} = \{\tau^{L_i} \mid L_i \in \mbox{SeedEnv}(n) \}
	\end{gather*}
	Using the simplified example from before, this would mean that
	$\mbox{SeedEnv}(3) = \{L_1, L_2, L_3\}$ and
	$\text{Demos} = \{\tau_1^{L_1}, \tau_2^{L_2}, \tau_3^{L_3}\}$. In the following, we refer
	to each $L_i \in \mbox{SeedEnv(n)}$ as \textit{seed level}.
	
	The reward function is learned via AIRL on the reduced SeedEnv
	environment instead of the fully-procedural ProcEnv. To distinguish
	expert from non-expert trajectories, the discriminator thus cannot rely on
	memorized level characteristics seen in expert demonstrations, but instead must
	consider the behavior represented by the state-action sequence of the
	trajectory.
	
	Once the discriminator is trained on SeedEnv, the learned reward
	function can be used to train a new agent on the full ProcEnv
	environment. The disentanglement property of AIRL encourages the reward function
	to be robust to the change of dynamics between different levels, assuming a
	minimum number of seed levels necessary to generalize across level
	configurations.

	In summary, we observe that there are two sources of discriminative features in
	expert trajectories: those related to the \emph{level}, and those related to
	\emph{agent behavior}. If AIRL is applied naively to PCG environments, the
	discriminator is prone to overfitting to level characteristics seen during
	expert demonstrations instead of focusing on the expert behavior itself. On the
	one hand, by reducing discriminator training to the SeedEnv -- the set
	of expert demonstration levels -- we force the discriminator to focus on
	trajectories and to avoid \textit{overfitting to level characteristics}. On the
	other hand, SeedEnv must contain enough levels to enable the resulting
	reward function to \textit{generalize beyond levels} in the reduced
	ProcEnv sample. We show empirically in the next section that the number
	of levels required to generalize beyond levels sampled in ProcEnv is much
	smaller than the number required to avoid overfitting, which may be infeasibly
	large for PCG environments with many configuration options.

	%%%%%%%%%%%%%%%%%%%%%%%%%%%%%%%%%%%%%%%%%%%%%%%%%%%%%%%%%%%%%%%%%%%%%%%%%%%%%%%%%%%%%%%%%%%%%%%%%%%%%%%%%%%%
	\section{Experimental results}
	\label{sec:experiments}
	
	We evaluate our method on two different PCG environments: Minigrid
	\citep{minigrid} and DeepCrawl \citep{sesto19}. For all experiments, we train an
	agent with Proximal Policy Optimization \citep{ppo} on the ground-truth,
	hard-coded reward function and then generate trajectories from this trained
	expert policy to use as demonstrations for IRL. The apprenticeship learning
	metric is used for IRL evaluation: agent performance is measured based on the
	ground-truth reward after having been trained on the learned IRL reward model.
	
	We use the state-only AIRL algorithm with all modifications described in section
	\ref{sec:modifications} to learn a reward function in all experiments. We
	also trained policies with state-only GAIL but, as it is not an IRL method,
	we cannot re-optimize the obtained model, so we instead transfer the learned
	policy from the SeedEnv to ProcEnv.
	
	To highlight the importance of our SeedEnv approach for learning good rewards in
	the context of PCG environments, we perform the following ablations for each
	task:
	\begin{itemize}[noitemsep,nolistsep,leftmargin=0.5cm]
		\item \textbf{DE-AIRL (ours)}: We train a reward function on SeedEnv and use it to train a PPO agent on ProcEnv. We show results for a varying number $n$ of seed levels in $\mbox{SeedEnv}(n)$.
		\item \textbf{AIRL without disentanglement}: We train a reward function on SeedEnv, but without the shaping term $\phi_\omega(s)$ which encourages robustness to level variation.
		\item \textbf{Naive AIRL}: We apply AIRL directly on ProcEnv and show results for a varying number $n$ of demonstrations.
		\item \textbf{GAIL}: We train a policy with GAIL on SeedEnv and then evaluate it on ProcEnv.
	\end{itemize}
	Details on network architectures and hyperparameters, 
	including an ablation study for the AIRL modifications described in section
	\ref{sec:modifications}, are given in Appendix \ref{sec:appendix_hypers} and 
	Appendix \ref{sec:appendix_modifications} 
	\footnote{
	Supplementary Material with code 
	%and appendices including an ablation study for the AIRL 
    %modifications described in section \ref{sec:modifications} 
    is available at \url{http://tiny.cc/de_airl}}.
	
	\begin{figure*}
		\begin{center}
			\scalebox{0.83}{
				\setlength\tabcolsep{-6pt}
				\begin{tabular}{cccc}
					\includegraphics[width=0.37\textwidth]{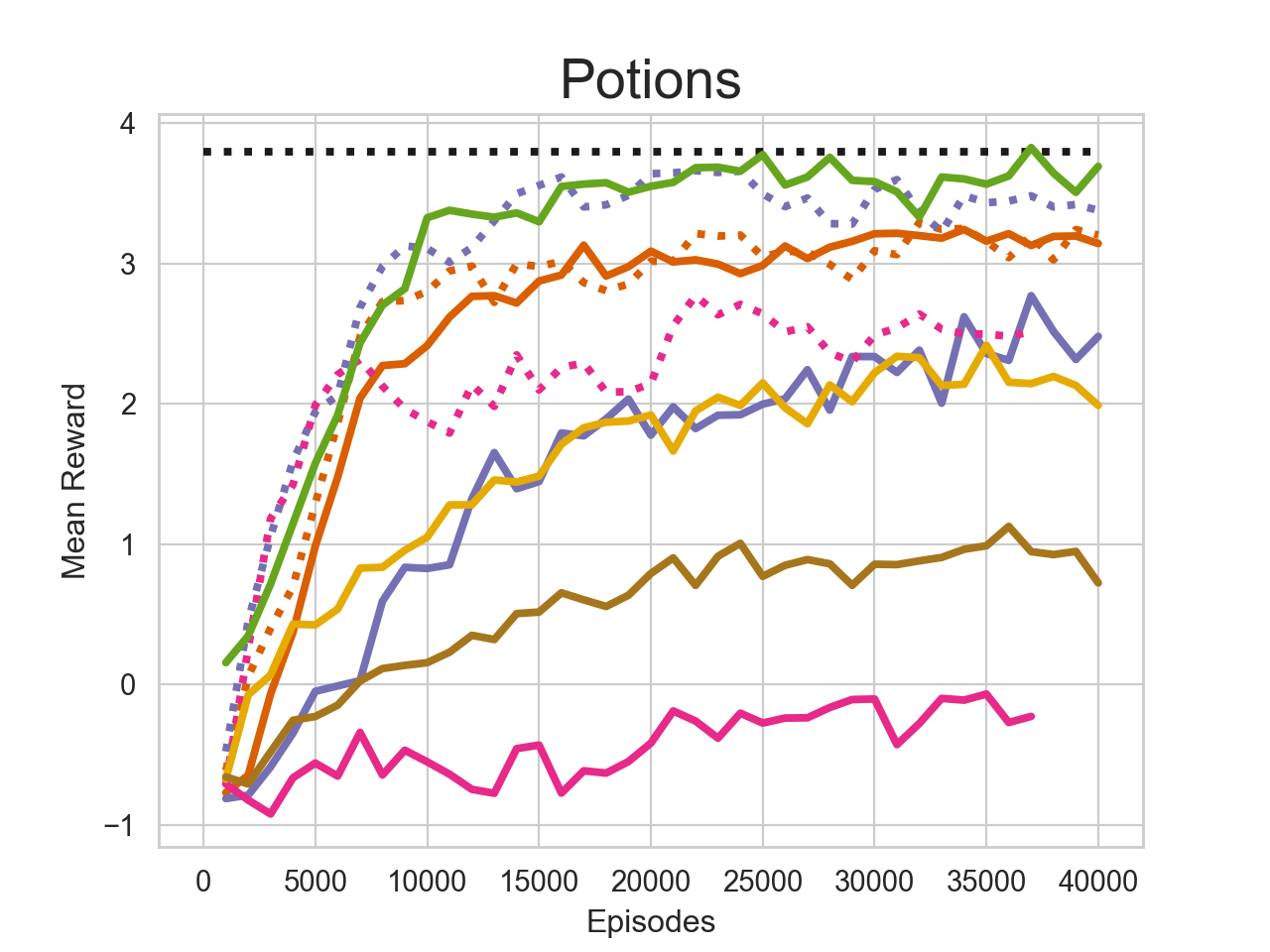} &
					\includegraphics[width=0.37\textwidth]{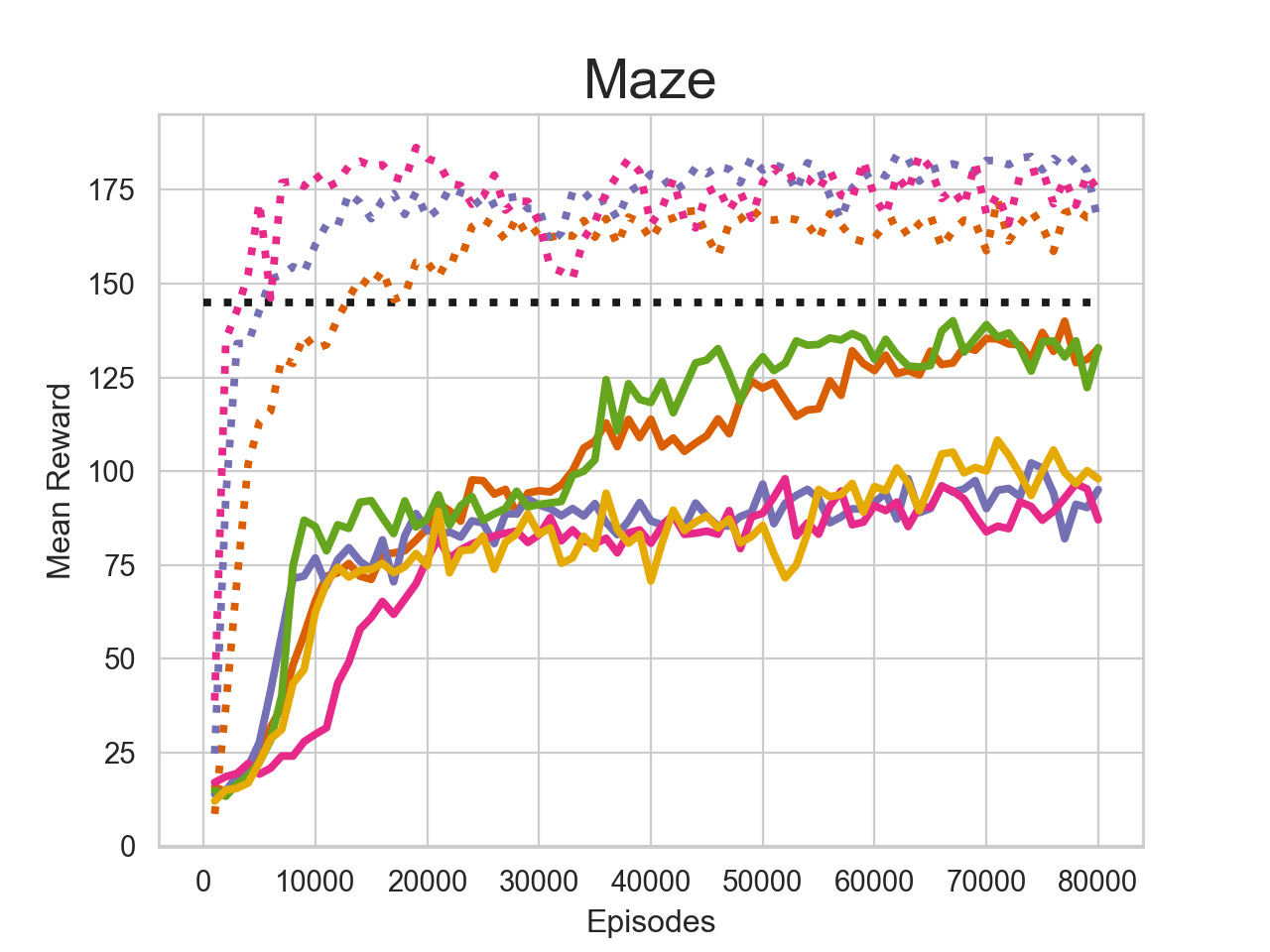} &
					\includegraphics[width=0.37\textwidth]{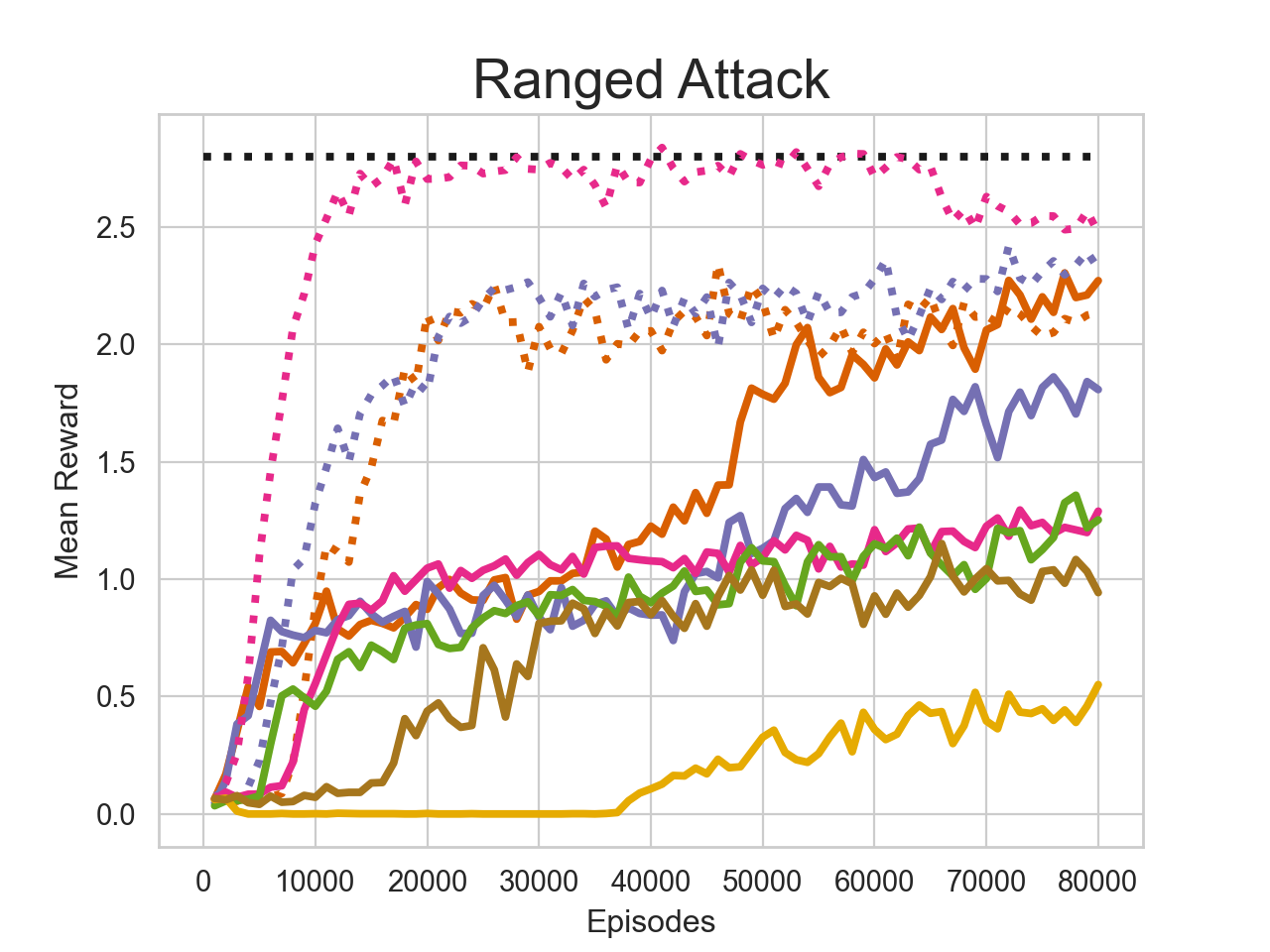} &
					\includegraphics[width=3.5cm]{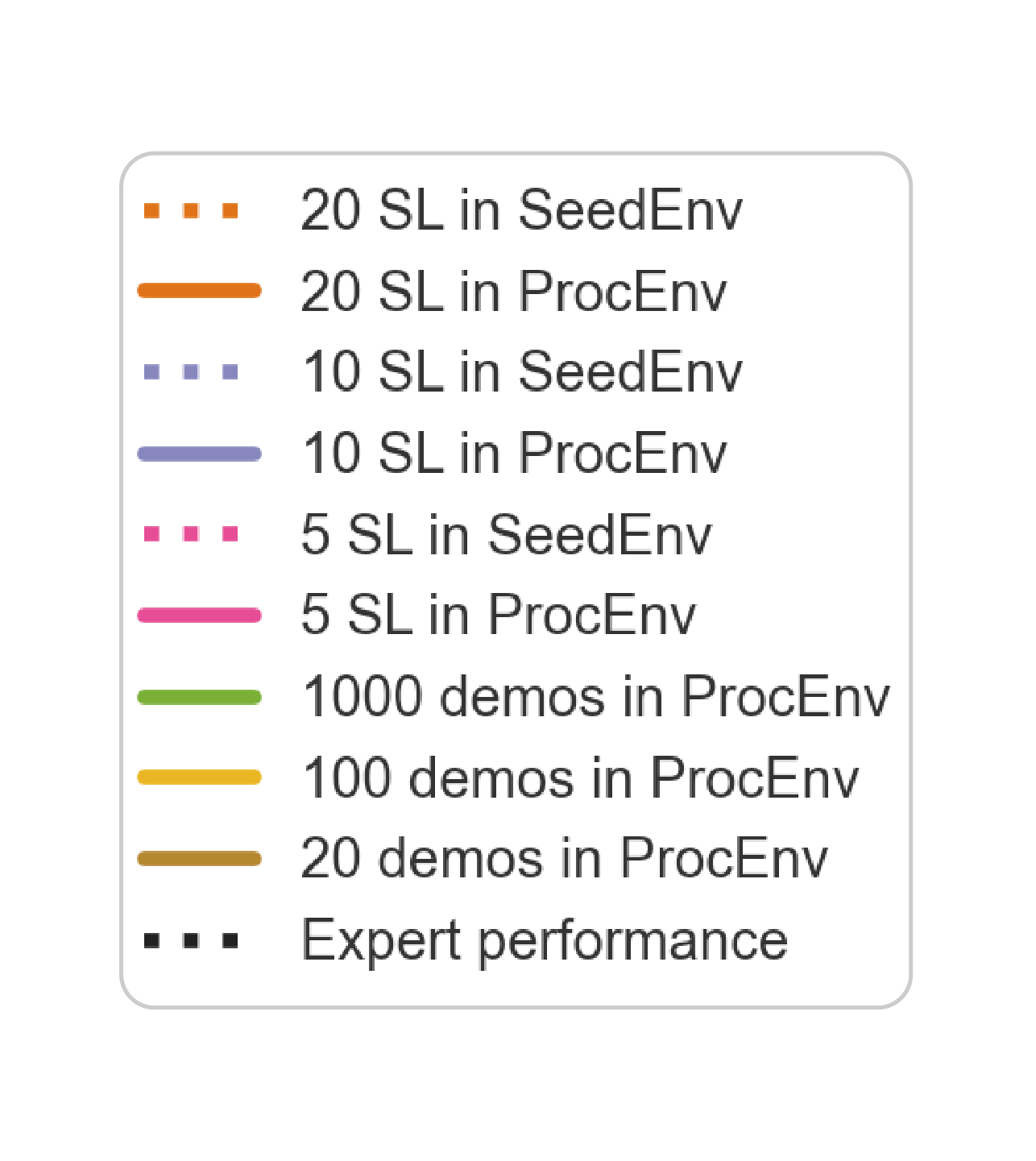} 
					\\
				\end{tabular}
			}
		\end{center}
		\caption{Experimental results on DeepCrawl tasks. Mean reward during training for: our DE-AIRL with different numbers of seed levels on both SeedEnv and ProcEnv, naive AIRL with different numbers of demonstrations on ProcEnv, and expert performance on ProcEnv.}
		\label{fig:deepcrawl_results}
	\end{figure*}
	\begin{figure*}
		\begin{center}
			\scalebox{0.73}{
				\setlength\tabcolsep{-6pt}
				\begin{tabular}{cccc}
					\includegraphics[width=0.37\textwidth]{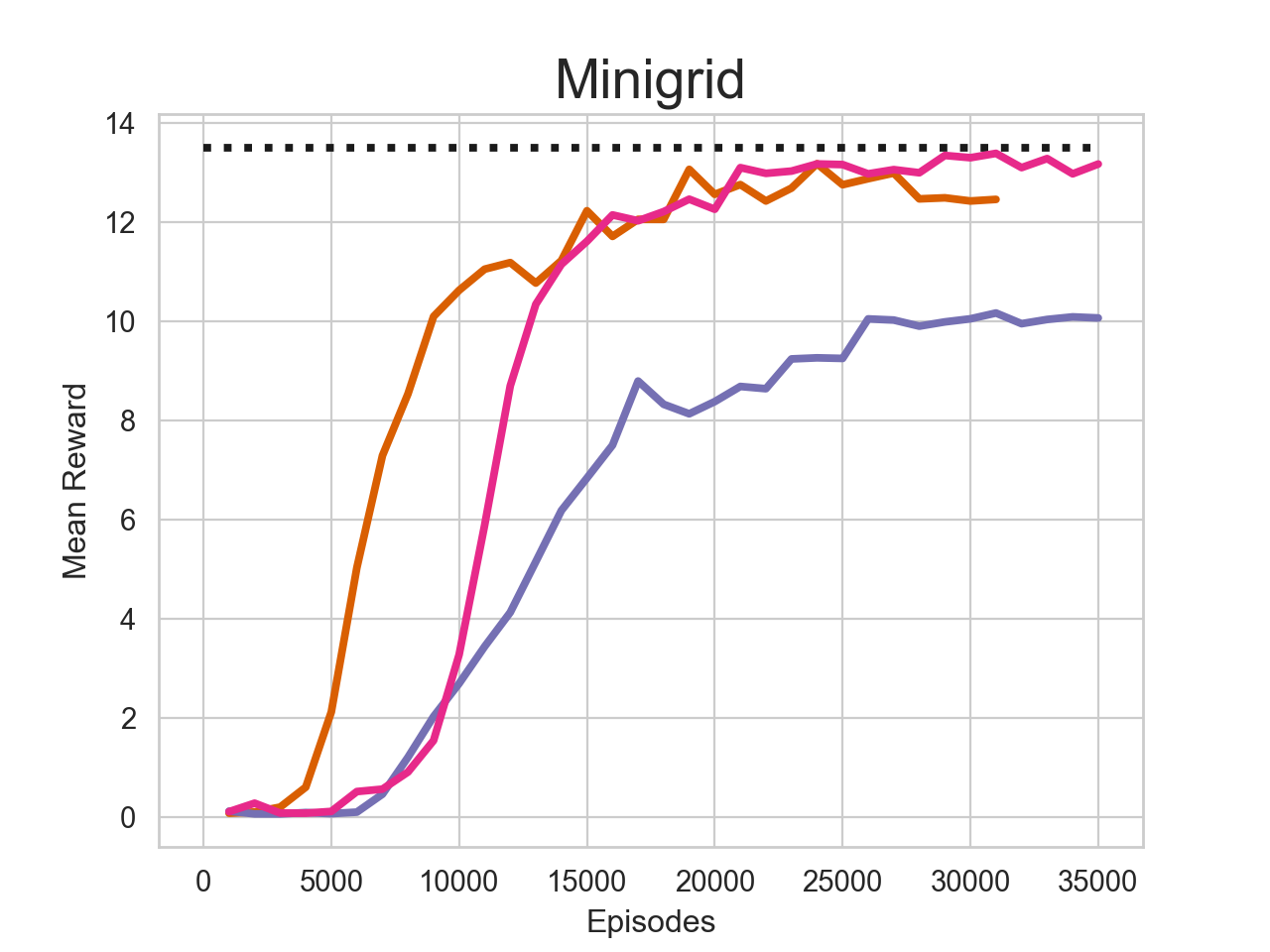} &
					\includegraphics[width=0.37\textwidth]{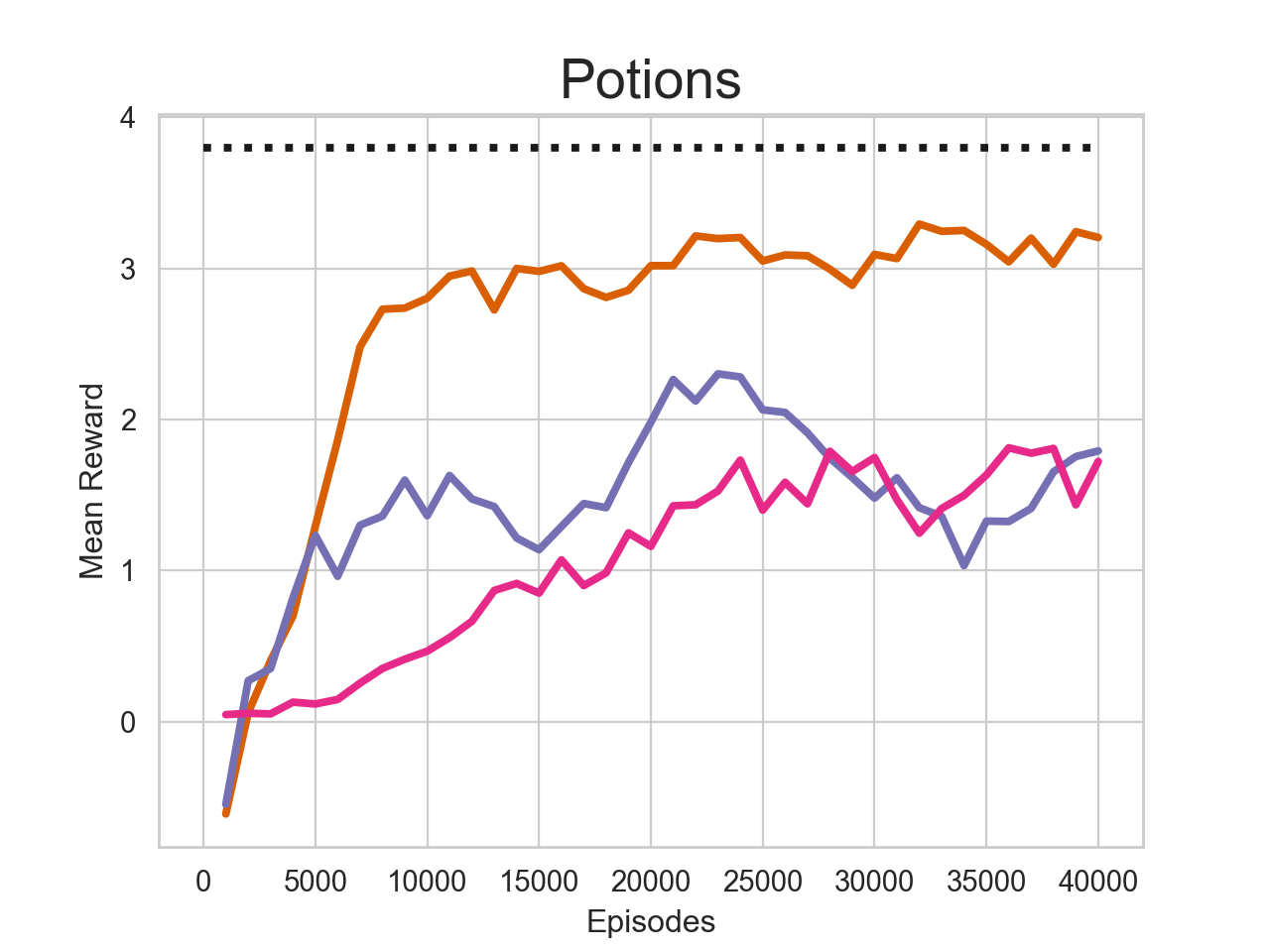} &
					\includegraphics[width=0.37\textwidth]{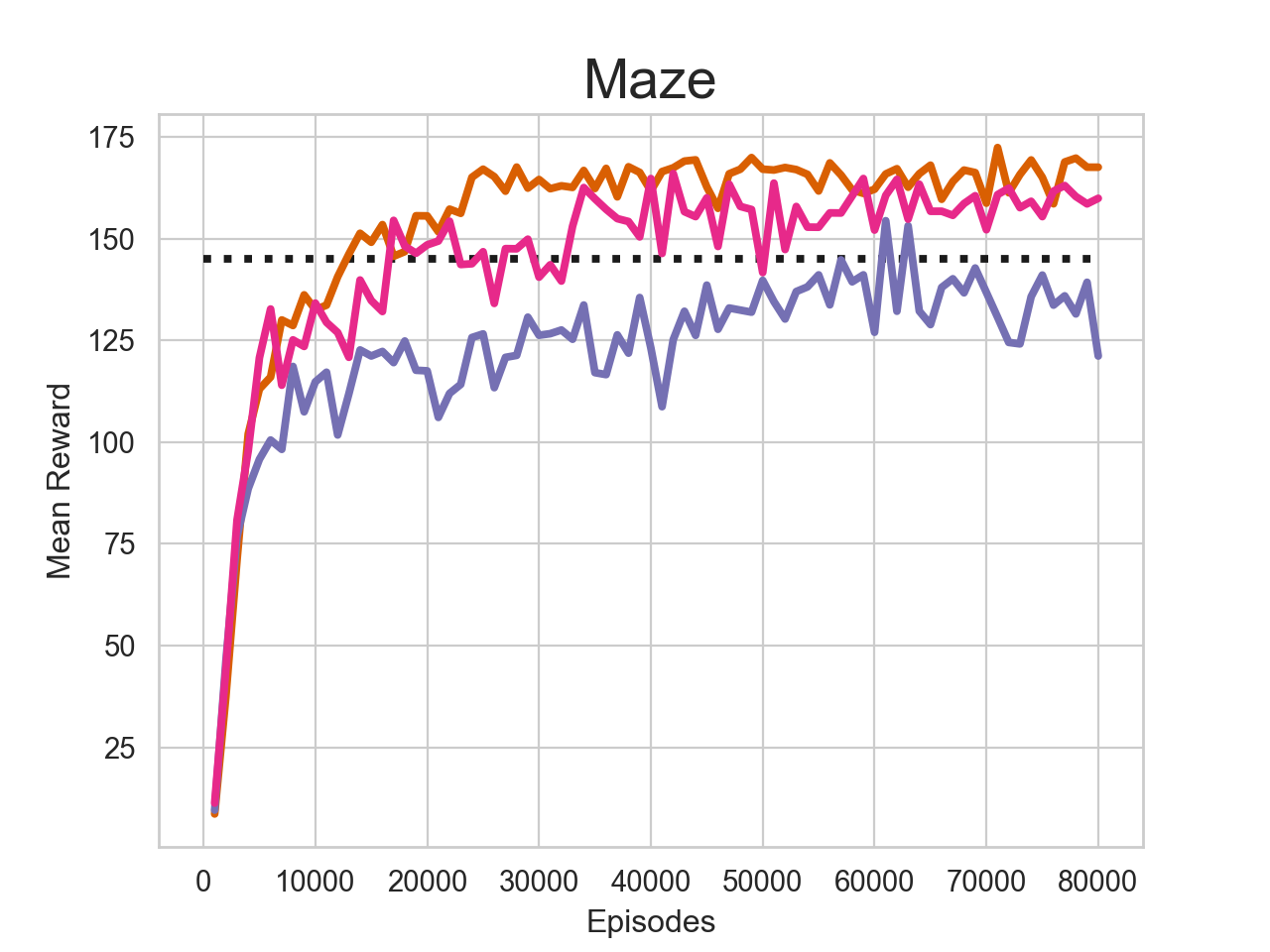} &
					\includegraphics[width=0.37\textwidth]{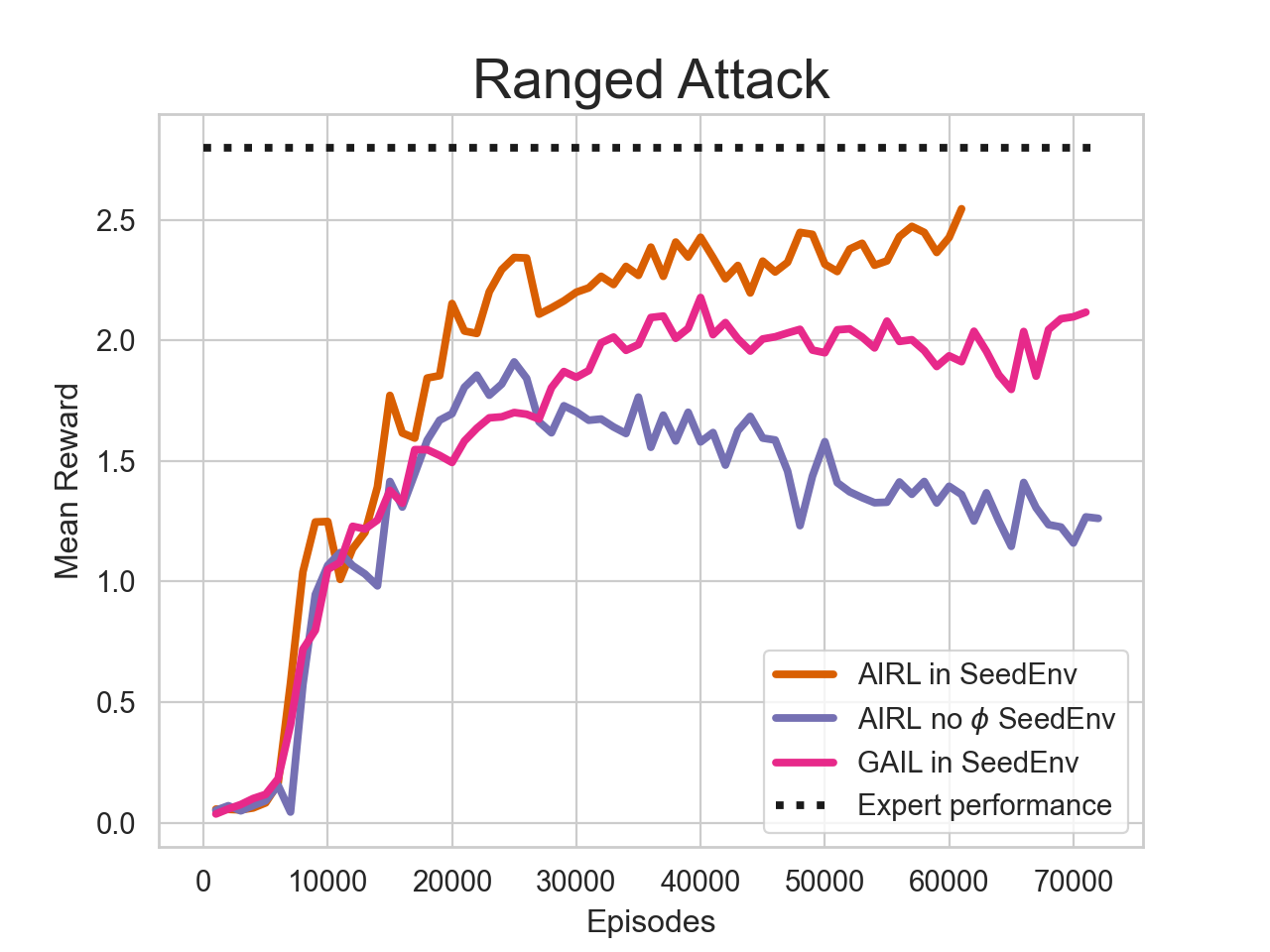} \\
				\end{tabular}
			}
		\end{center}
		\caption{Mean reward on SeedEnv throughout training for AIRL without the
			shaping term, and for GAIL, plus expert performance on ProcEnv for
			comparison. SeedEnvs consist of 40 levels for Minigrid, and 20 levels for
			the DeepCrawl tasks.}
		\label{fig:values_results}
	\end{figure*}
	
	\subsection{Performance on Minigrid}
	\label{minigrid}
	Minigrid is a grid world environment with multiple variants. For our
	experiments, we use the \textit{MultiRoom} task: a PCG environment consisting of
	a $15 \times 15$ grid, where each tile can contain either the agent, a door, a
	wall, or the goal. See figure \ref{fig:screen} for an example screenshot of the
	environment. The aim of the agent is to explore the level and arrive at
	the goal tile by navigating through 2 or 3 rooms connected via doors.
	The shape and position of the rooms, as well as the position of the goal and
	the initial location of the agent, are random. Each episode lasts a maximum of 30
	steps. The ground-truth reward function gives $+1.0$ for each step the agent
	stays at the goal location. The action space consist of 4 discrete actions:
	\textit{move forward}, \textit{turn left}, \textit{turn right} and \textit{open
		door}.
	
	As the results in figure \ref{fig:minigrid_results} show, using a
	SeedEnv with only 40 levels and the associated 40 demonstrations, AIRL
	is able to extrapolate a good reward function enabling the agent to achieve
	near-expert performance in ProcEnv. However, if we train a reward model
	with only 40 demonstrations directly on the full PCG environment, we obtain an
	inadequate reward function and consequently a poor agent policy. This is also
	demonstrated by the loss curves: the loss of the discriminator with 40
	demonstrations on ProcEnv converges to zero very quickly, indicating
	the \textit{overfitting to level characteristics} we discussed in section
	\ref{sec:seedmethod}. The results also show that 40 is a good number of seed
	levels for SeedEnv: whereas we find a good policy for SeedEnv
	with only 30 seed levels, the reward function does not generalize beyond the
	expert levels to be useful on ProcEnv. Moreover, the plots show that
	naive AIRL is not successful on ProcEnv with even 100 -- so more than
	twice as many -- expert trajectories. Only with 1000 demonstrations does naive
	AIRL achieve near-expert performance, showing that our DE-AIRL is much
	more demonstration-efficient.
	
	\subsection{Performance on DeepCrawl}
	\label{sec:deepcrawl}
	
	DeepCrawl is a Roguelike game built for studying the applicability of DRL
	techniques in video game development. The visible environment at any time is a
	grid of $10 \times 10$ tiles. Each tile can contain the agent, an opponent, an
	impassable object, or collectible loot. The structure of the map and object
	locations are procedurally generated at the beginning of each episode.
	Collectible loot and actors have attributes whose values are randomly chosen as
	well. The action space consists of 8 movement actions: horizontal, vertical and
	diagonal.
	
	For our experiments, we use three different tasks defined in the DeepCrawl
	environment (see figure~\ref{fig:screen}):
	\begin{itemize}[noitemsep,nolistsep,leftmargin=0.5cm]
		\item \textbf{Potions}: The agent must collect red potions while avoiding all
		other collectible objects. The ground-truth reward function gives $+1.0$ for
		collecting a red potion and $-0.5$ for collecting any other item. An episode
		ends within 20 steps.
		
		\item \textbf{Maze}: In this variant, the agent must reach a randomly located
		goal in an environment with many impassable obstacles forming a maze. The goal
		is a static enemy and there are no collectible objects. The reward function
		gives $+10.0$ for each step the agent stays in proximity to the goal. Episodes
		end after 20 timesteps.
		
		\item \textbf{Ranged Attack}: For this task, the agent has two additional
		actions: melee attack and ranged attack. The goal of the agent is to hit a
		static enemy with only ranged attacks until the enemy is defeated. The
		ground-truth reward function gives $+1.0$ for each ranged attack made by the
		agent. The levels are the same as for the Potions task, plus a randomly located
		enemy. Episodes end after 20 timesteps.
	\end{itemize}
	
	Even for the more complex DeepCrawl tasks, the results in figure
	\ref{fig:deepcrawl_results} show that our demonstration-efficient AIRL
	approach allows agents to learn a near-expert policy for ProcEnv with
	few demonstrations: in two of the three tasks only 20 demonstrations
	are necessary, while for the Ranged Attack task 10 already
	suffice. Similar to Minigrid, the naive AIRL approach directly applied
	on ProcEnv does not achieve good performance even with 100
	demonstrations -- so with more than five times as many
	demonstrations. With 1000 demonstrations, naive AIRL reaches similar
	performance on Potions and Maze, but still not on the Ranged Attack
	task. In figure 2 of the Supplementary Material, we provide more
	detailed results for the DeepCrawl experiments, including the
	evolution of discriminator losses which behave consistently with what we have observed for the Minigrid environment.
	
	\subsection{Importance of disentanglement}
	We claimed above that the use of a disentangling IRL algorithm like AIRL is
	fundamental for PCG games. We test this experimentally by training an
	AIRL reward function without the shaping term $\phi(s)$ on a SeedEnv. As the
	plots in figure \ref{fig:values_results} show, this modified version does not
	achieve the same level of performance as the full disentangling AIRL on all
	tasks. We believe this is due to the variability of levels in SeedEnv: removing
	$\phi(s)$ takes away the disentanglement property, which results in the reward
	function no longer being able to generalize, even for the small set of fixed
	seed levels. Similar results were observed by \citeauthor{airlvsgail}.
	
	We also train a state-only GAIL model on a SeedEnv. On Minigrid and Maze the
	policy reaches near-expert performance, while on Potions and Ranged Attack it
	resembles the performance of AIRL without $\phi(s)$. We believe that this
	discrepancy is caused by the different degree of \textit{``procedurality''} of
	these tasks: for Potions and Ranged Attack, there are many different collectible
	objects with procedural parameters -- in fact, all entities and their attributes
	are chosen randomly at the beginning of each episode. For the other two tasks,
	the number of procedural choices is smaller, consisting only of the static
	obstacles and no attributes. The degree of procedurality presumably allows GAIL
	to achieve good results on SeedEnv for Minigrid and Maze, but not for Potions
	and Ranged Attack. However, on none of the tasks does GAIL reach the level of
	performance of our demonstration-efficient AIRL approach when transferring
	policies from SeedEnv to ProcEnv, as shown in table \ref{tab:transfer}. Note
	that, as we have mentioned before, GAIL is not an IRL method and hence cannot be
	re-optimized on the ProcEnv environment, contrary to AIRL, so this shortcoming
	is not unexpected.
	
	\renewcommand{\arraystretch}{1.3}
	\begin{table*}
		\caption{Average ground-truth episode reward over 100 episodes on ProcEnv. Our
			AIRL approach trains an agent directly on ProcEnv using the reward model
			learned on SeedEnv, whereas this is not possible for GAIL, hence the GAIL
			policy is trained on SeedEnv and then transferred to ProcEnv.}
		\label{tab:transfer}
		\begin{center}
				\scalebox{0.85}{
					\begin{tabular}{c||c | c || c | c c c}
						\toprule
						& \multicolumn{2}{c||}{Minigrid} & \multicolumn{4}{c}{DeepCrawl} \\
						& Seed levels & MultiRoom & Seed levels & Potions & Maze & Ranged Attack\\
						\midrule
						DE-AIRL (ours) & 40 & \textbf{12.19} & 20 & \textbf{3.78} & \textbf{141.87} & \textbf{2.30} \\
						% Seed (policy transfer)  & 40 & 9.04  & 20 & 2.80 & 53.86 & 0.70\\
						\hline
						\multirow{2}{*}{GAIL} & 40 & 9.00  & 20 & 1.71 & 33.77 & 1.01\\
						& 100 & 9.21  & 100 & 2.38 & 66.00 & 1.11\\
						\bottomrule
				\end{tabular}}
		\end{center}
	\end{table*}
	\renewcommand{\arraystretch}{1.0}
	
	\section{Conclusion}
	\label{discussion}
	We have presented an IRL approach, DE-AIRL, which is based on AIRL with a few modifications to stabilize performance, and is able to find a good reward function for PCG
	environments with only few demonstrations. Our method introduces a SeedEnv which consists of only a few levels
	sampled from the PCG level distribution, and which is used to train the reward model instead of the full fully-procedural environment. In doing so, the learned reward model is able to generalize beyond the SeedEnv levels to the fully-procedural
	environment, while it simultaneously avoids overfitting to the expert demonstration levels. We have shown that DE-AIRL substantially reduces the number of required expert demonstrations as compared to AIRL when directly applied on
	the PCG environment. Moreover, the experiments illustrated that the success of our approach derives from the disentanglement
	property of the reward function extrapolated by AIRL. Finally, we compared to an imitation learning approach, GAIL, and observed that DE-AIRL generalizes better than the GAIL policy when
	transferring from the expert demonstration levels to the fully-procedural environment.
	
	A disadvantage of our method is that we do not know the required number of seed
	levels prior to training. In this direction, an interesting next step
	would be to understand what minimum number of seed levels is required to obtain a good reward function as well as a good policy. For instance, starting
	with a small number of seed levels, how can we choose additional seed levels optimally based on the training and learned reward function so far?
	
\bibliography{main}

\newpage

\appendix
\section{Implementation details}
\label{sec:appendix_hypers}
In this section we give additional details on the network architectures used for
DE-AIRL on the Minigrid and DeepCrawl environments.

\subsection{Network Structures}
\citeauthor{airl} use a multilayer perceptron for reward and policy models, however we
use Convolutional Neural Networks (CNNs) like \citeauthor{adam}. Moreover, we use
Proximal Policy Optimization (PPO) \citep{ppo} instead of Trust-Region Policy
Optimization (TRPO) \citep{trpo} as in the original paper.
\begin{itemize}
    
\item{\textbf{Minigrid.}
The policy architecture consists of two branches. The first branch takes the
global view of the $15 \times 15$ grid, and each tile is represented by a
categorical value that describes the type of element in that tile. This input is
fed to an embedding layer and then to a convolutional layer with $3 \times 3$
filters and 32 channels. The second branch is like the first, but receives as
input the $7 \times 7$ categorical local view of what the agent sees in front of
it. The outputs of the convolutional layers are flattened and concatenated
together before being passed through a fully-connected layer of size 256. The
last layer is a fully connected layer of size 4 that represents the probability
distribution over actions.}

The reward model and the shaping term $\phi_\omega$ have the same architecture.
Unlike the policy network, they take only the global categorical map and pass it
through an embedding layer, two convolutional layers with $3 \times 3$ filters
and 32 channels followed by a maxpool, and then two fully-connected layers of
size 32 and a final fully-connected layer with a single output. All other layers
except the last one use leaky-ReLu activations.

\item{\textbf{Potions and Maze.}
The convolutional structure of the policy of the Potions and Maze tasks are the same of
\citeauthor{sesto19} without the \textit{``property module''} and the LSTM layer. The
reward model takes as input only the global view, then it is followed by a
convolutional layer with $1 \times 1$ filters and size 32, by two convolutional
layers with $3 \times 3$ filters and 32 filters, two fully-connected layers of size
32, and a final fully-connected layer with a single output and no activation. The
shaping term $\phi_\omega$ shares the same architecture. We used leaky ReLu
instead of simple ReLu as used in DCGAN \citep{dcgan}.}

\item{\textbf{Ranged Attacks.}
In this case the policy has the complete structure of \citeauthor{sesto19} without
LSTM, and the reward model is the same of the previous tasks with the addition
of other two input branches that take as input two lists of properties of the
agent and the enemy. Both are followed by embedding layers and two fully
connected layers of size 32. The resulting outputs are concatenated together with
the flattened result of the convolutional layer of the first branch. This vector
is then passed to the same 3 fully connected layers of the potion task. The
shaping term shares the same architecture.}

\end{itemize}

\subsection{Hyperparameters}
In table \ref{tab:hyper} we detail the hyperparameters used for all tasks for
both policy and reward optimization.

\renewcommand{\arraystretch}{1.3}
\begin{table}
  \caption{Hyper-parameters for all the tasks. Most of the values were chosen after many
    preliminary experiments made with different configurations }
  \label{tab:hyper}
  \begin{center}
    \begin{small}
    \scalebox{0.85}{
      \begin{tabular}{c||c c c c}
        \toprule
        Parameter & Minigrid & Potions & Maze & Ranged Attack \\
        \midrule
        $\mbox{lr}_{policy}$	    & $5e^{-5}$ & $5e^{-5}$ & $5e^{-6}$ & $5e^{-5}$ \\
        $\mbox{lr}_{reward}$        & $5e^{-6}$ & $5e^{-4}$ & $5e^{-4}$ & $5e^{-4}$ \\
        $\mbox{lr}_{baseline}$	    & $5e^{-4}$ & $5e^{-4}$ & $5e^{-4}$ & $5e^{-4}$ \\
        entropy coefficient 		& $0.5$ & $0.1$ & $0.1$ & $0.1$ \\
        exploration rate 			& $0.5$ & $0.2$ & $0.2$ & $0.2$ \\
        $K$                         & $3$ & $3$ & $5$ & $3$ \\
        $\gamma$					& $0.9$ & $0.9$ & $0.9$ & $0.9$ \\
        max timesteps                   & $30$ & $20$ & $20$ & $20$ \\
        $\mbox{std}_{reward}$       & $0.05$ & $0.05$ & $0.05$ & $0.05$ \\
        \bottomrule
      \end{tabular}}
    \end{small}
  \end{center}
\end{table}

\begin{figure*}[ht!]
  \begin{center}
    \setlength\tabcolsep{-6pt}
    \begin{tabular}{cccc}
      \multicolumn{4}{c}{SeedEnv} \\
      \includegraphics[width=0.28\textwidth]{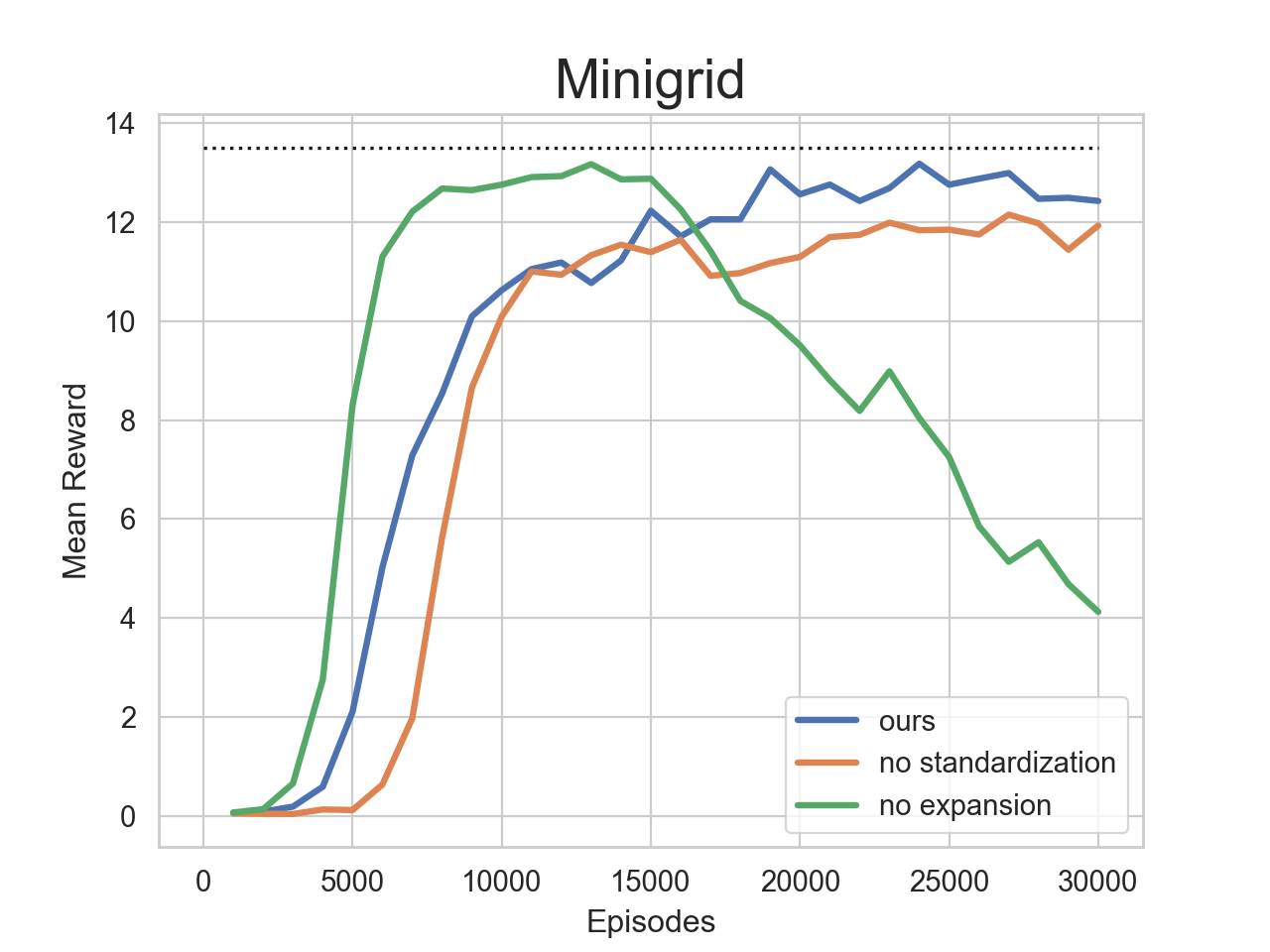} &
      \includegraphics[width=0.28\textwidth]{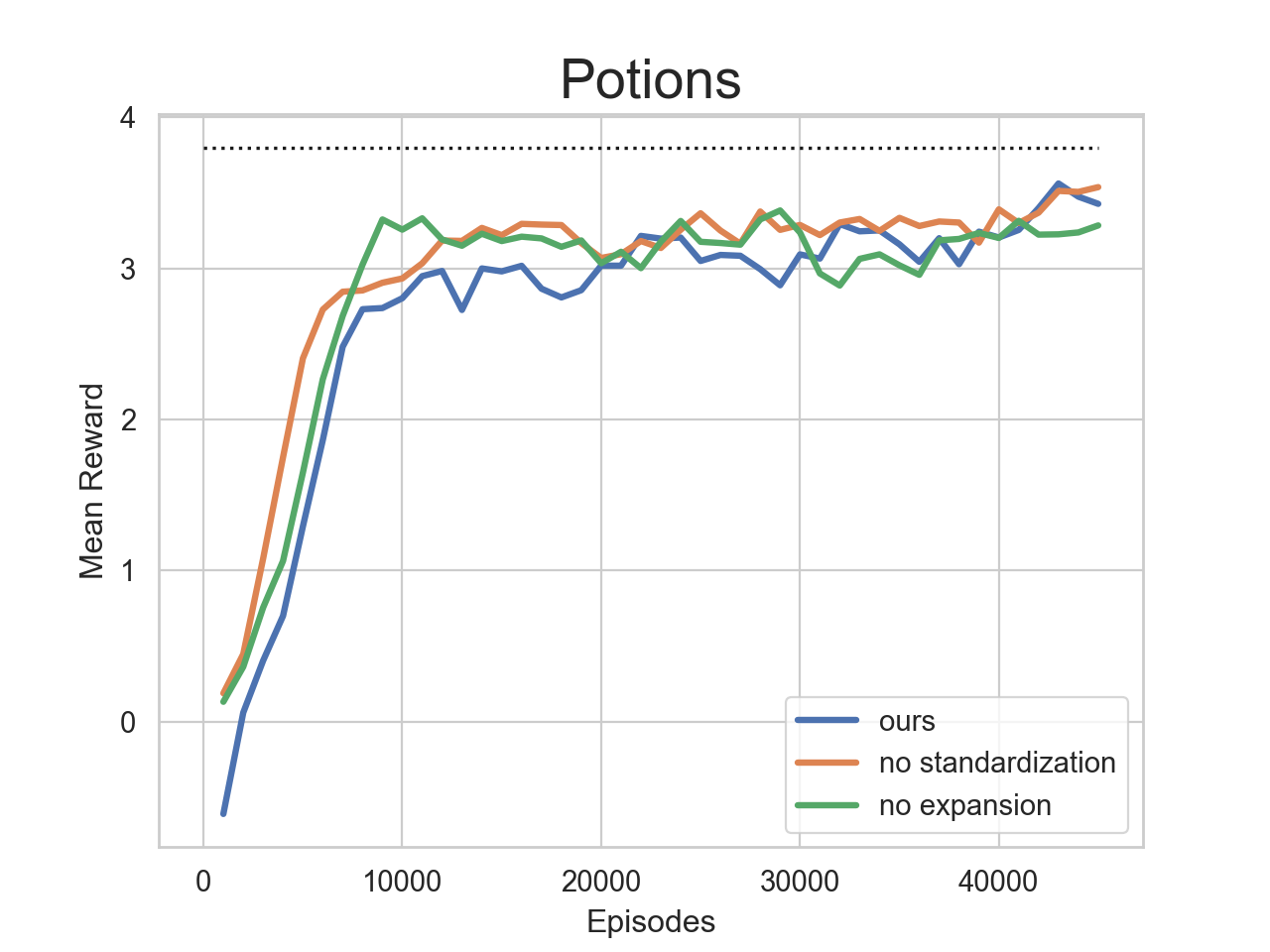} &
      \includegraphics[width=0.28\textwidth]{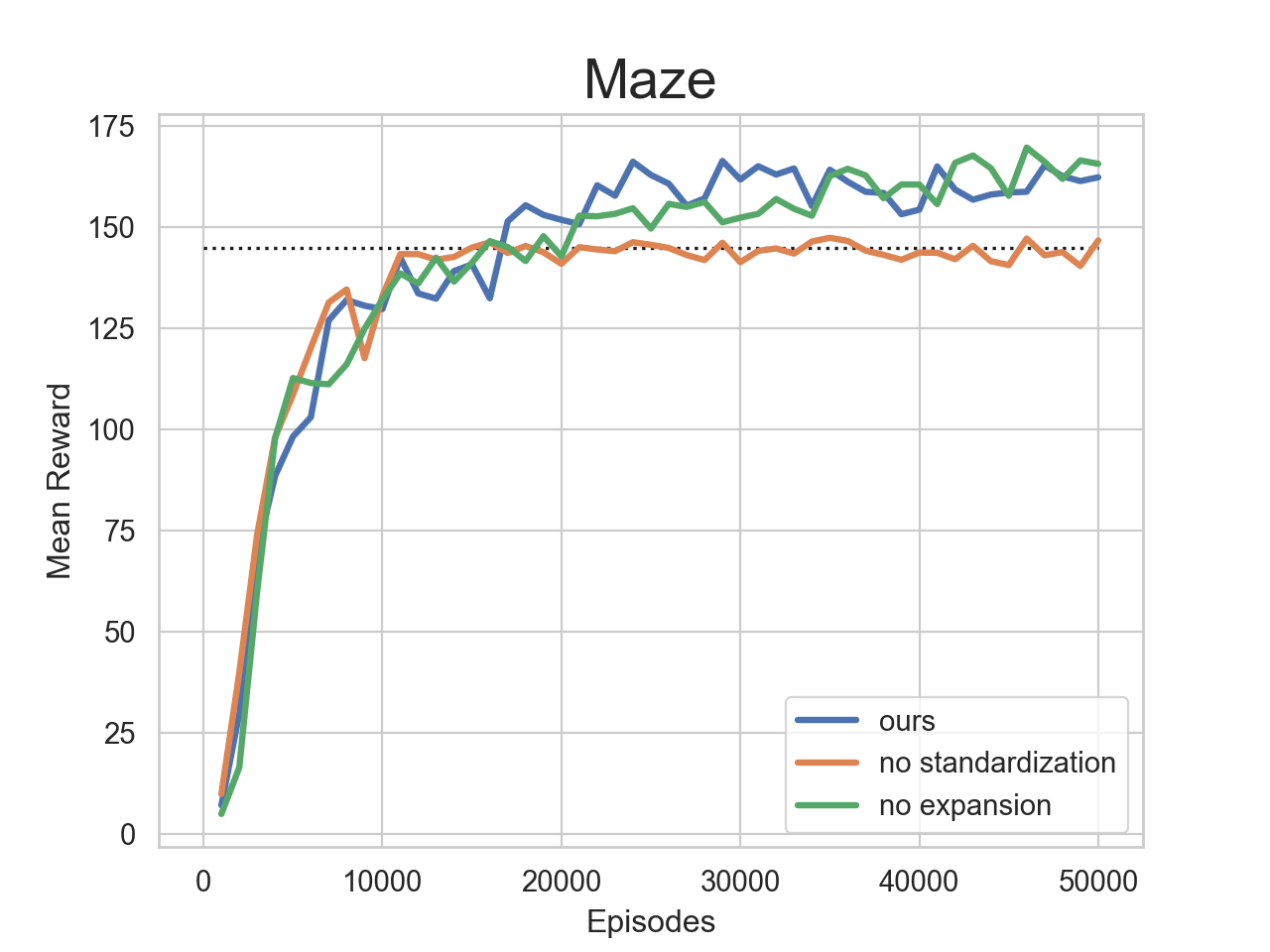} &
      \includegraphics[width=0.28\textwidth]{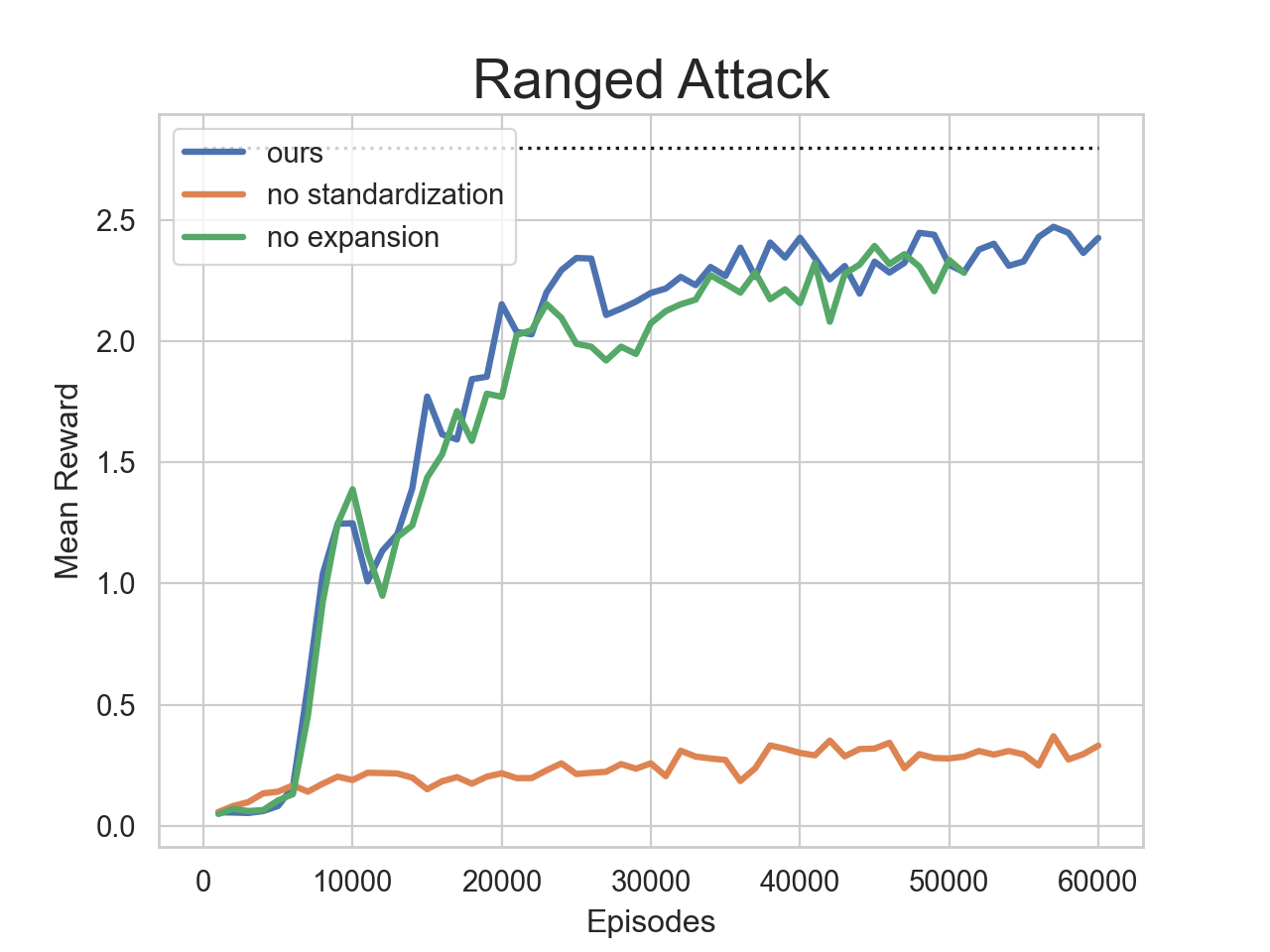}\\
      \multicolumn{4}{c}{ProcEnv} \\
      \includegraphics[width=0.28\textwidth]{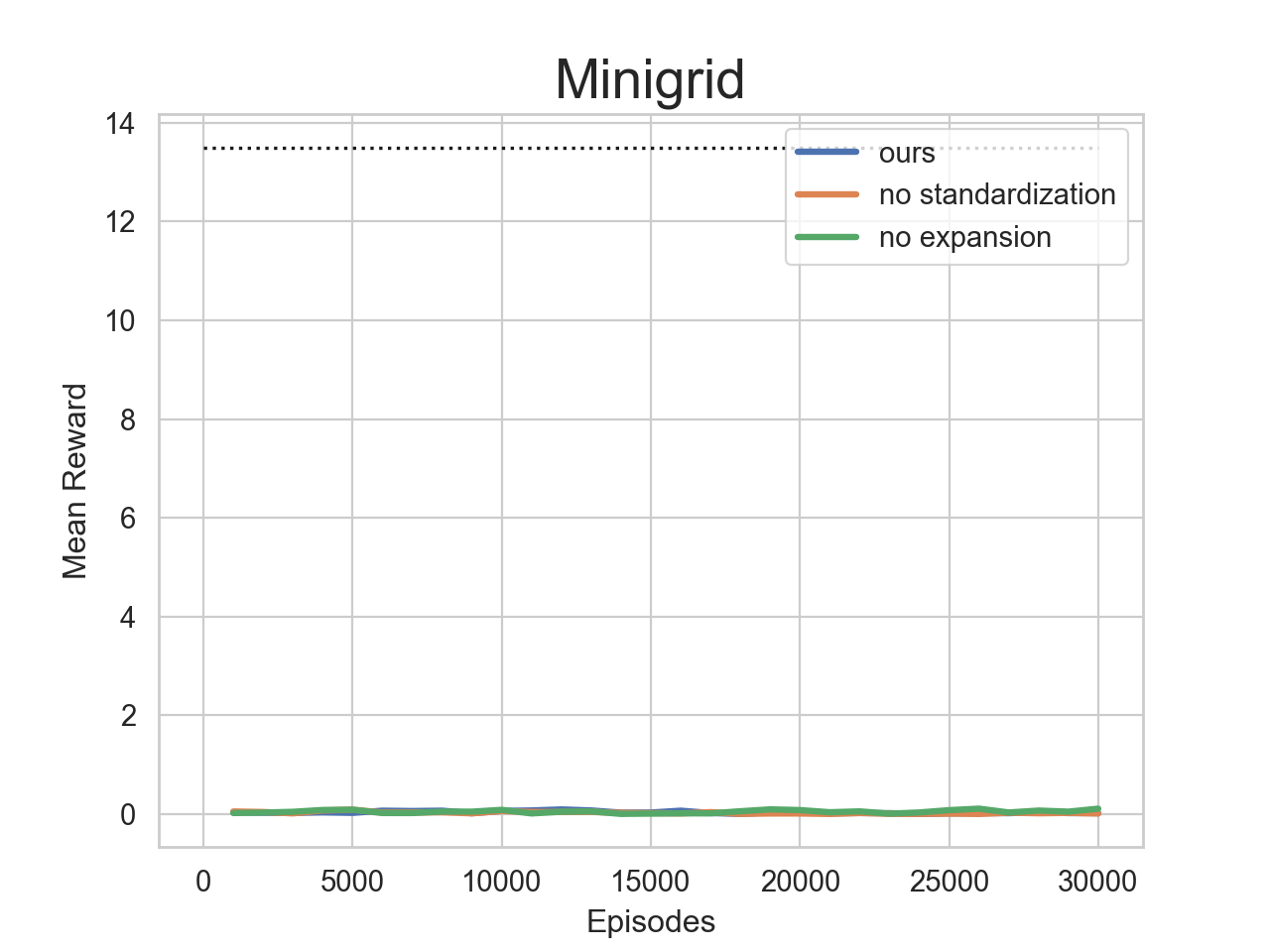} &
      \includegraphics[width=0.28\textwidth]{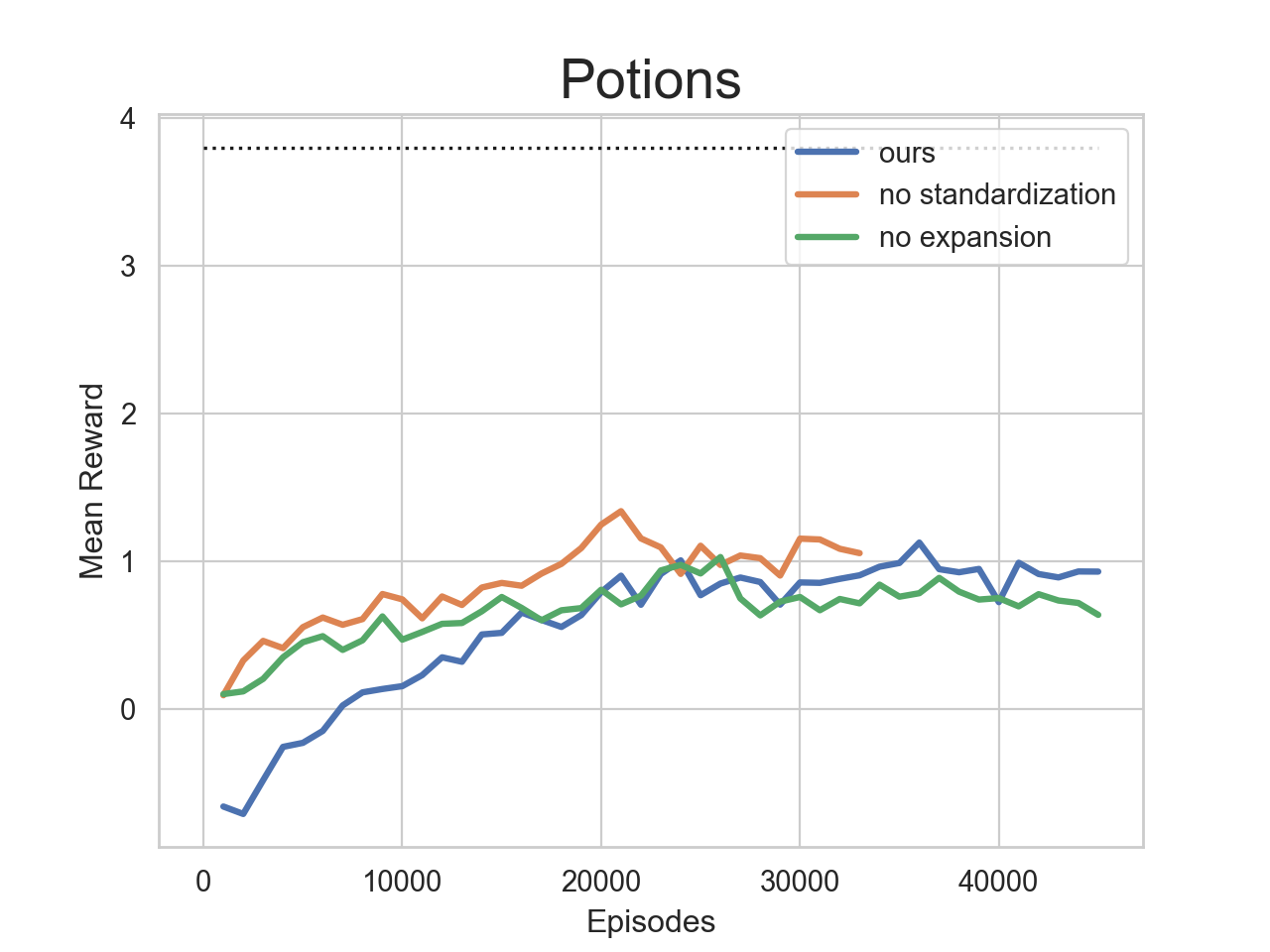} &
      \includegraphics[width=0.28\textwidth]{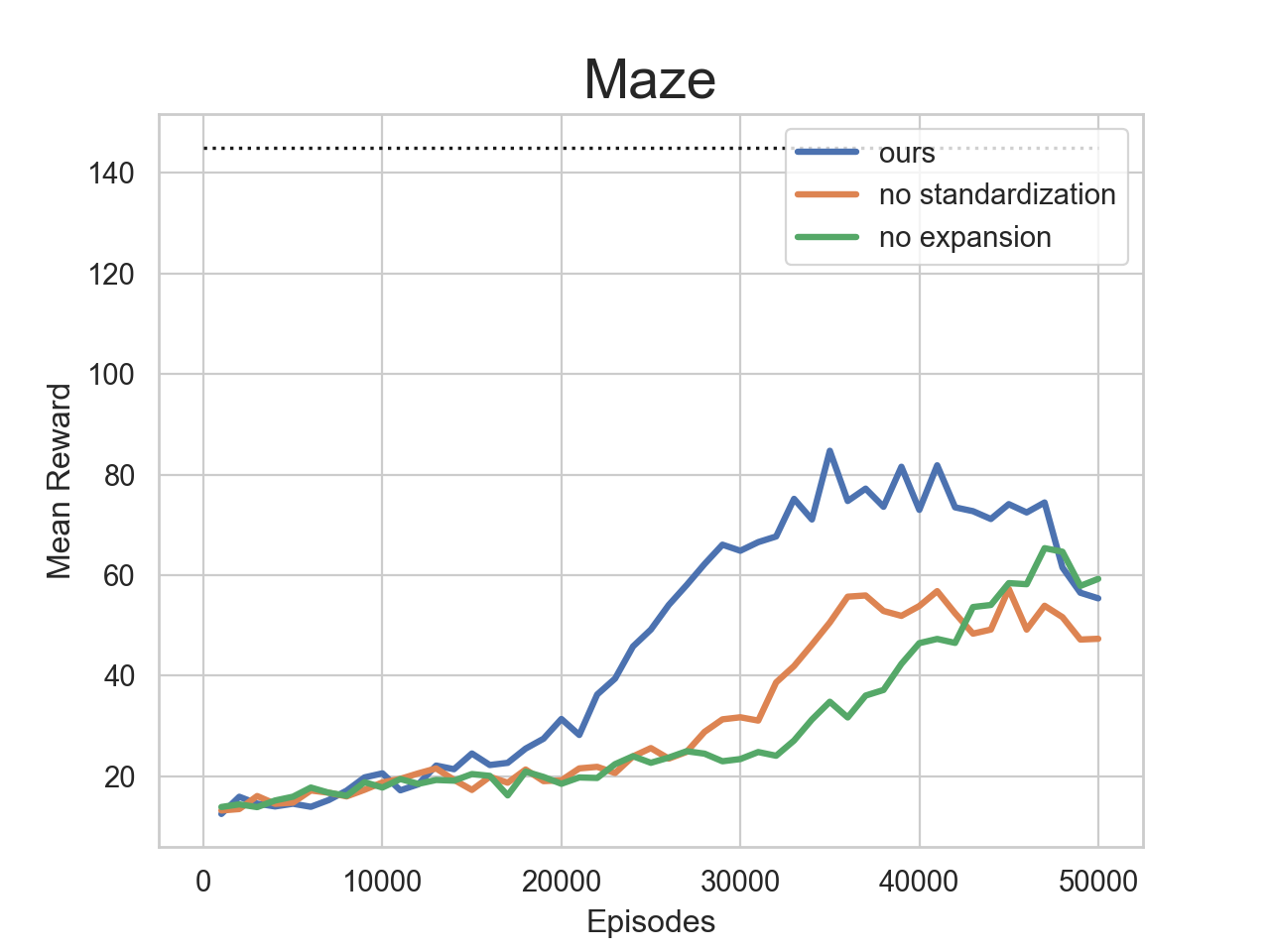} &
      \includegraphics[width=0.28\textwidth]{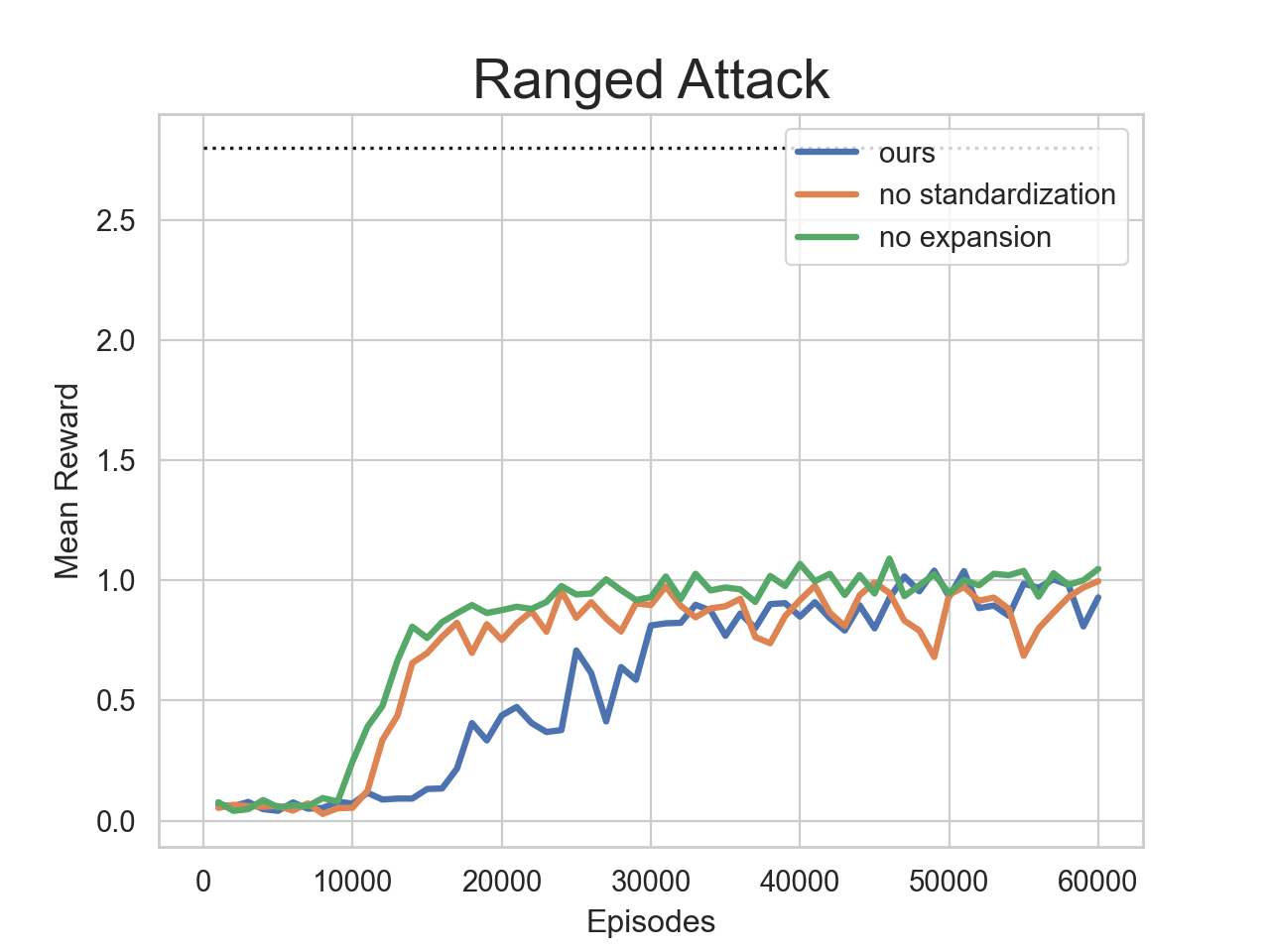} \\
	\end{tabular}
  \end{center}
  \caption{Ablation study of the modifications described in section 4 of the
    main text. The first row represents training in a SeedEnv, while the last
    row represents training in a ProcEnv. For all the DeepCrawl tasks we used 20
    seed levels and 20 demonstrations, while for Minigrid we used 40 seed levels
    and 40 demonstrations.}
  \label{fig:modifications}
\end{figure*}

\section{Effects of modifications to AIRL}
\label{sec:appendix_modifications}
In figure \ref{fig:modifications} we give an ablation study on both SeedEnv and
ProcEnv for the modifications to AIRL proposed in section 4 of the main text. The
plots show how the use of both reward standardization and policy dataset
expansion yield more stable and better results for the majority of the tasks on
both the environment types.

\section{Additional experimental results}
\begin{figure*}[ht!]
  \begin{center}
    \scalebox{1.0}{
      \setlength\tabcolsep{-6pt}
      \begin{tabular}{ccc}
        \includegraphics[width=0.37\textwidth]{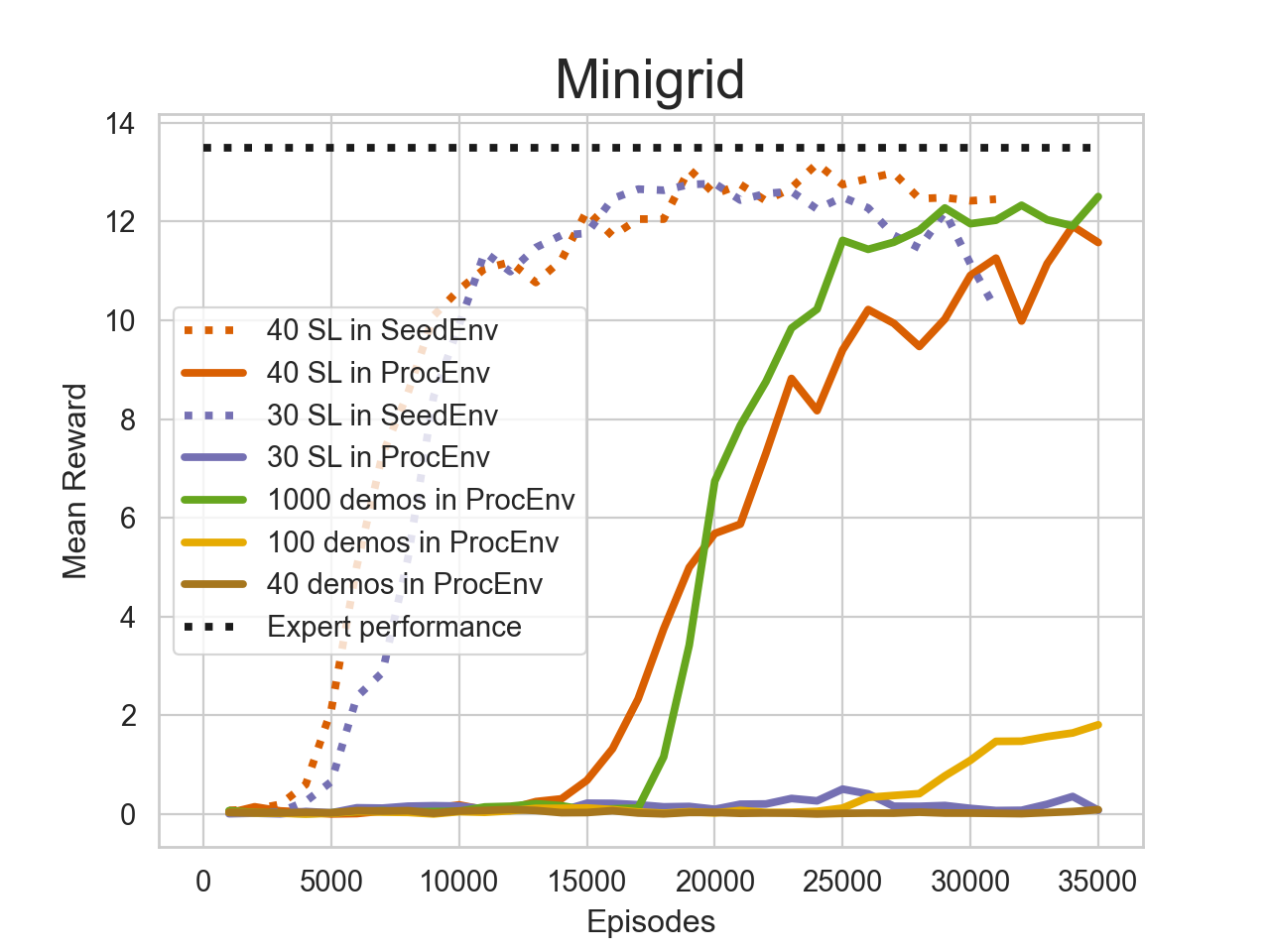} &
		\includegraphics[width=0.37\textwidth]{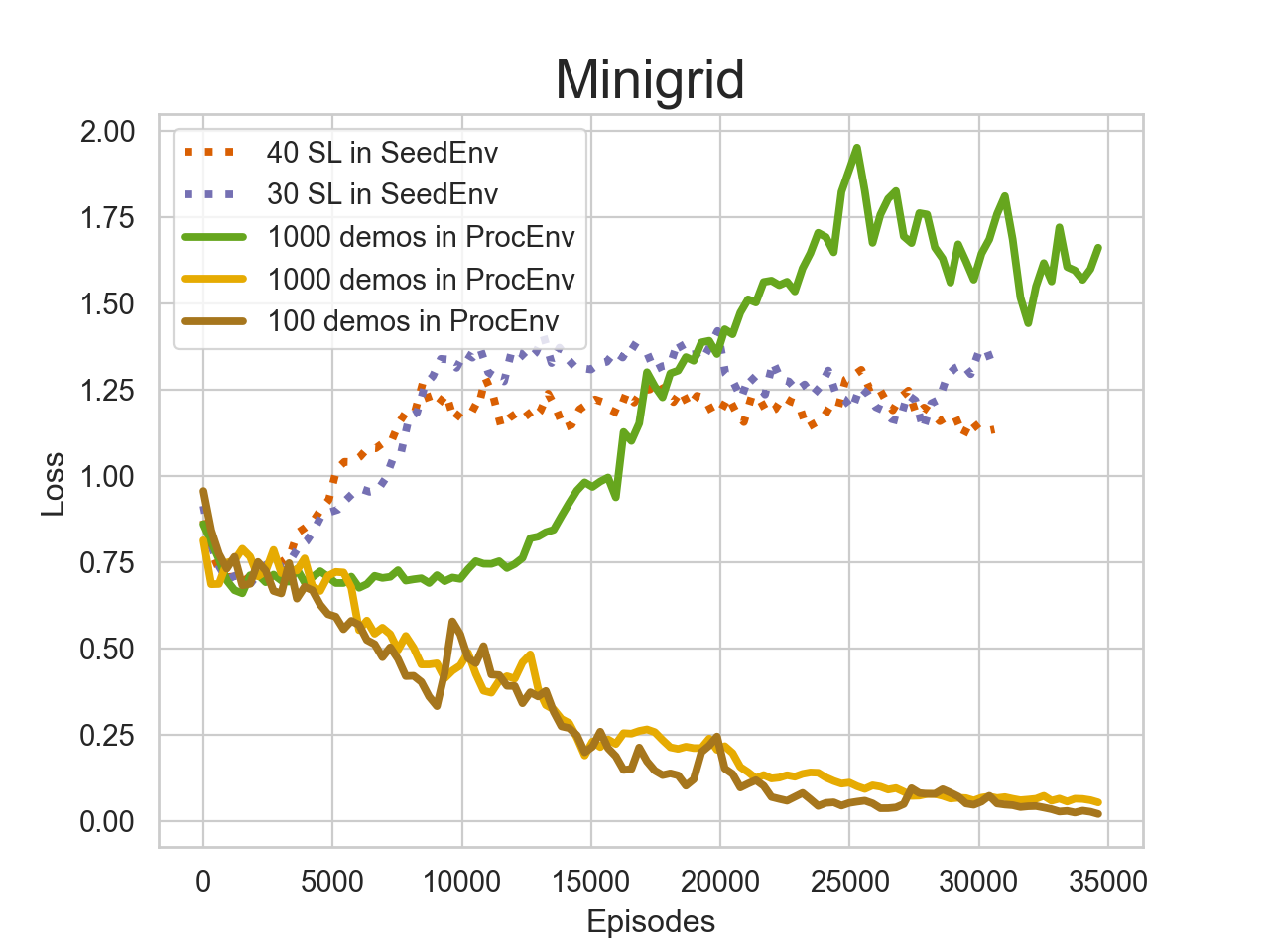} &
		\includegraphics[width=0.37\textwidth]{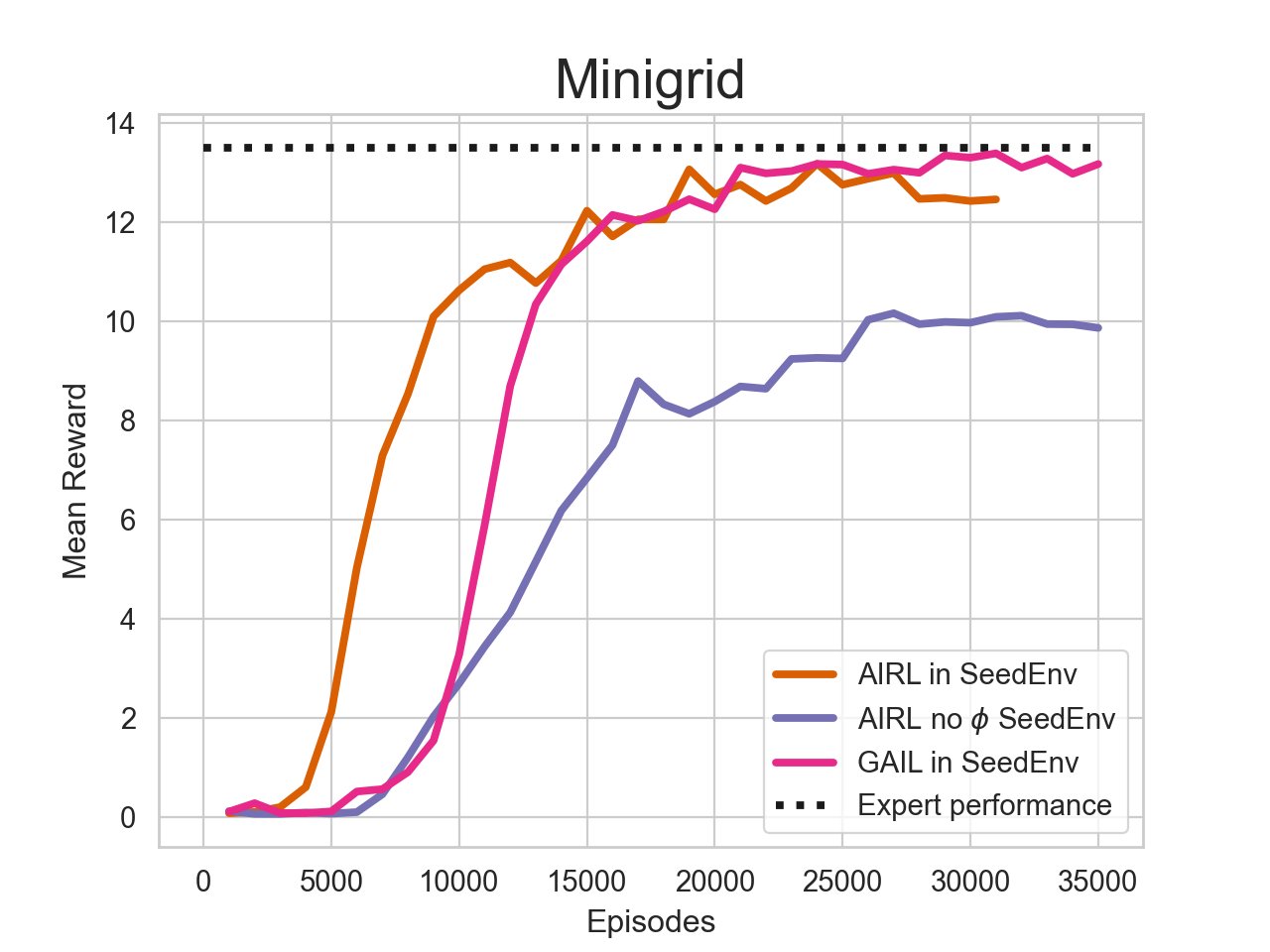} \\

		\includegraphics[width=0.37\textwidth]{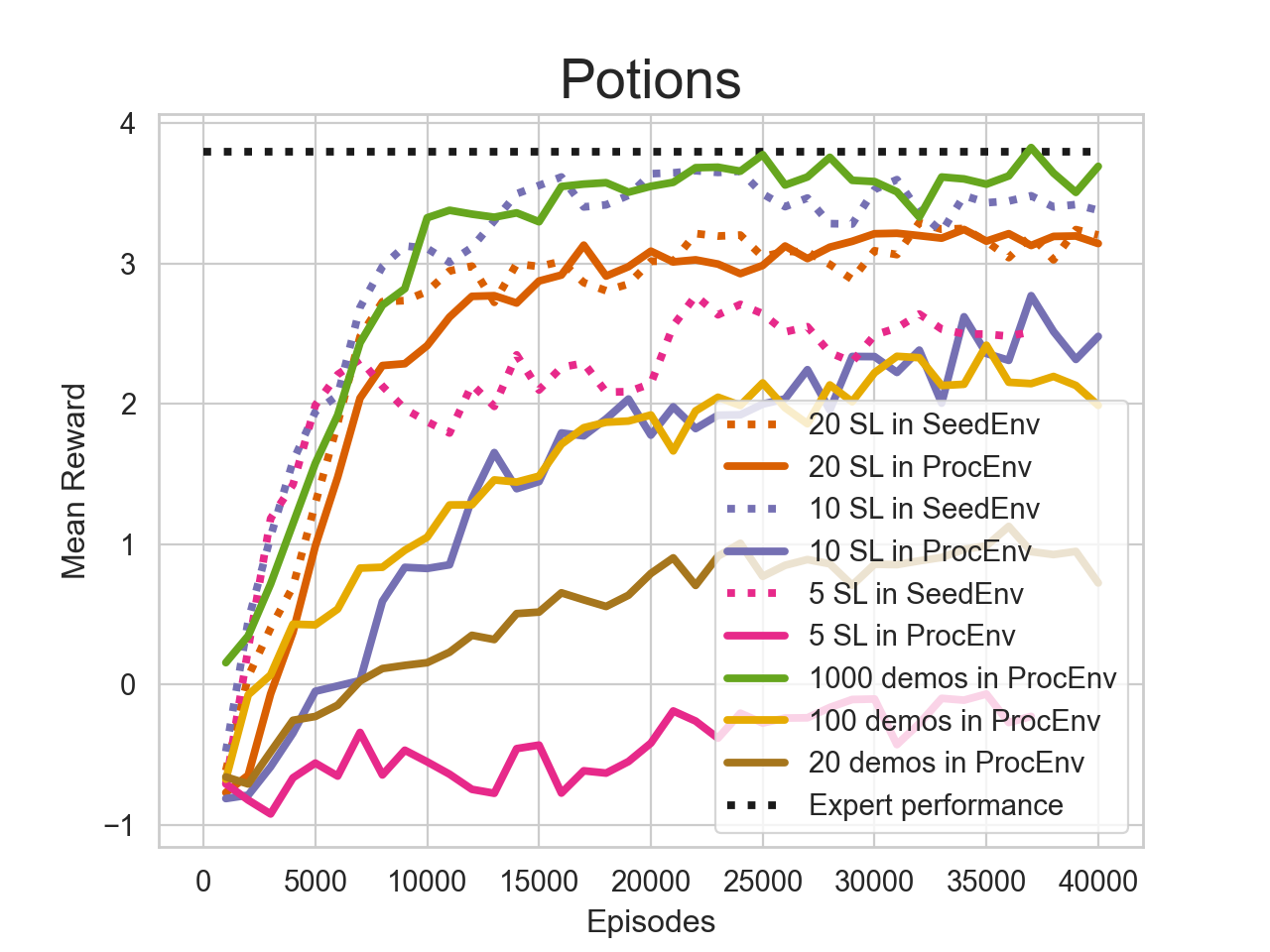} &
		\includegraphics[width=0.37\textwidth]{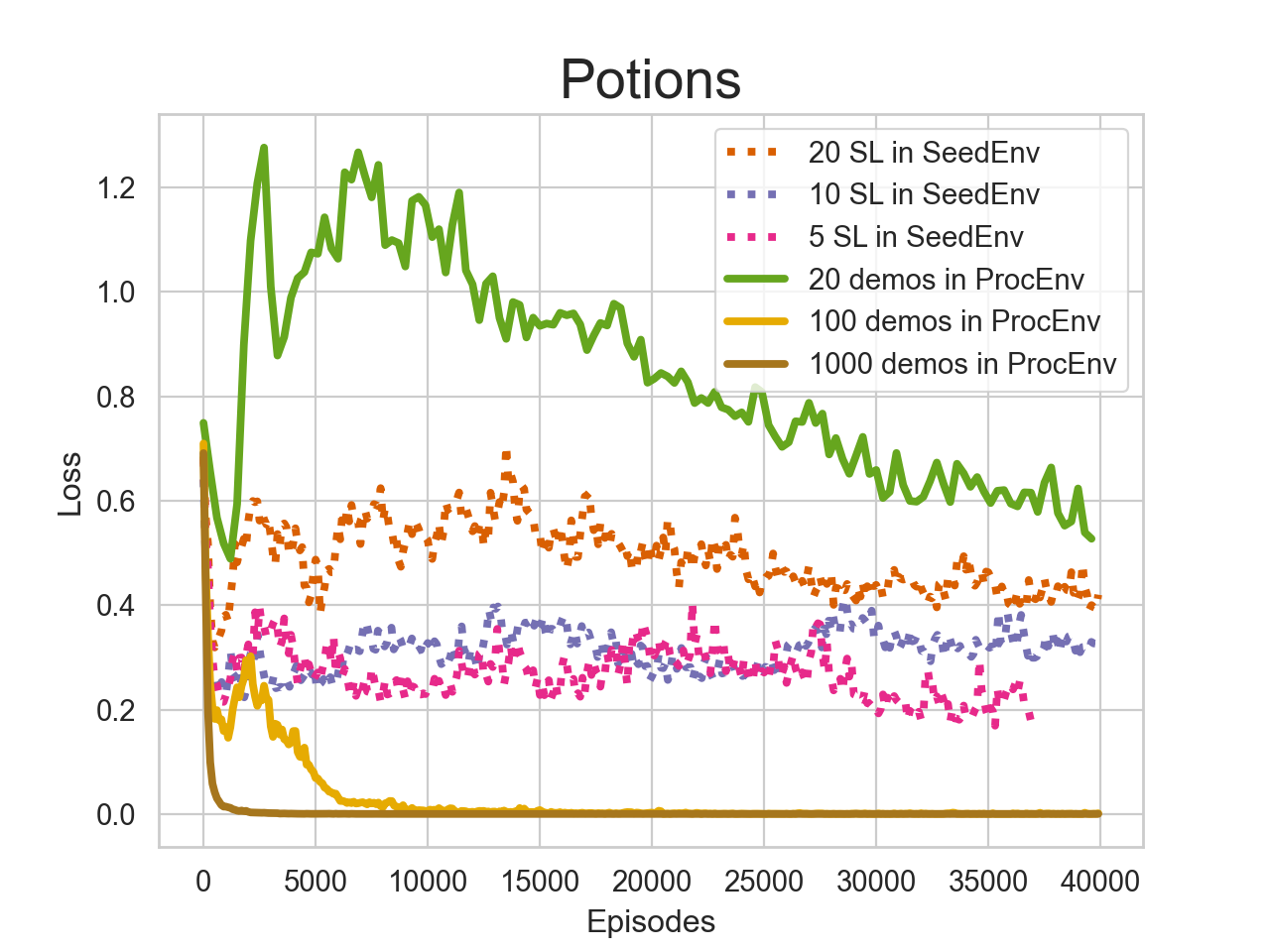} &
		\includegraphics[width=0.37\textwidth]{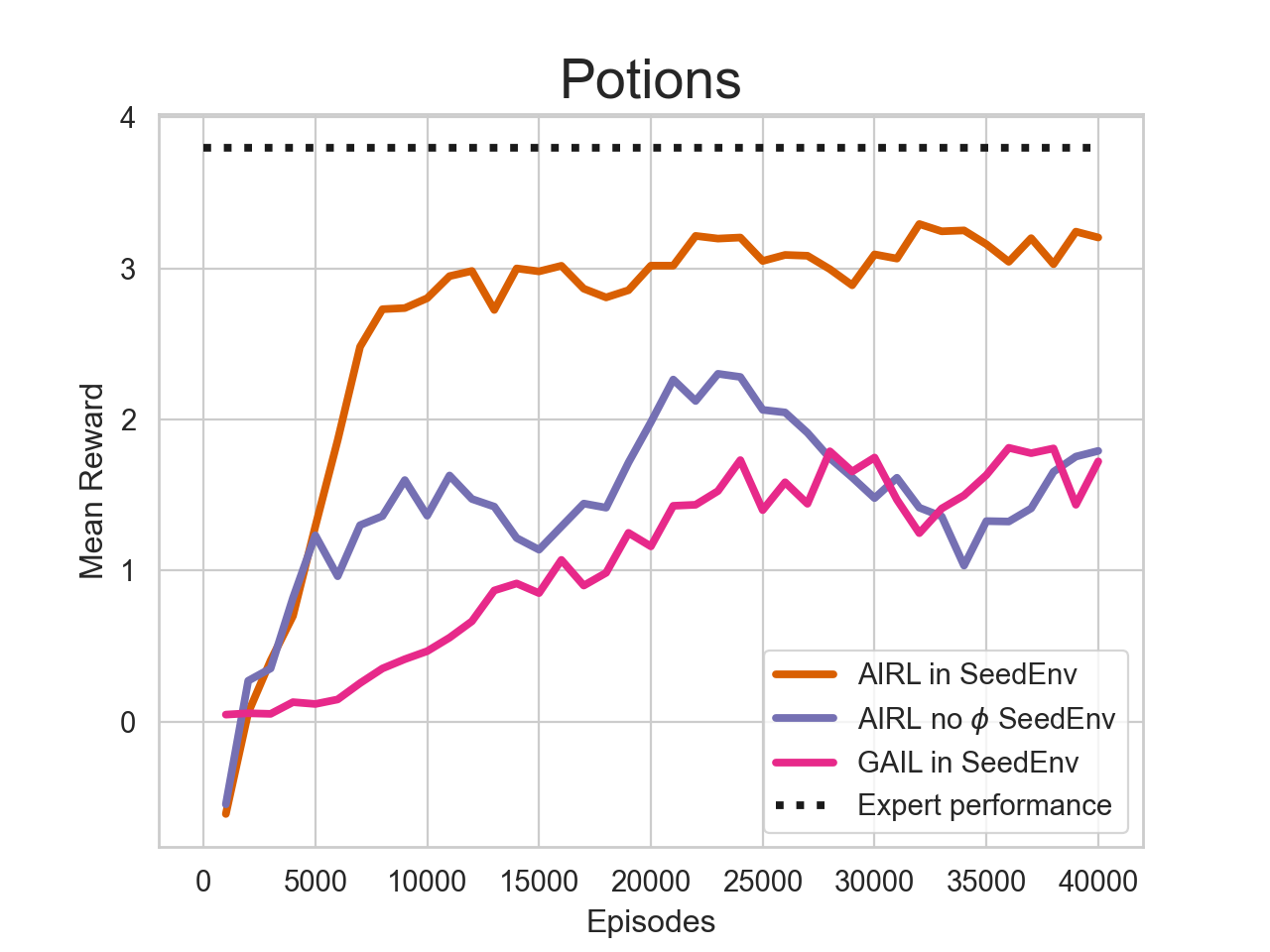} \\

		\includegraphics[width=0.37\textwidth]{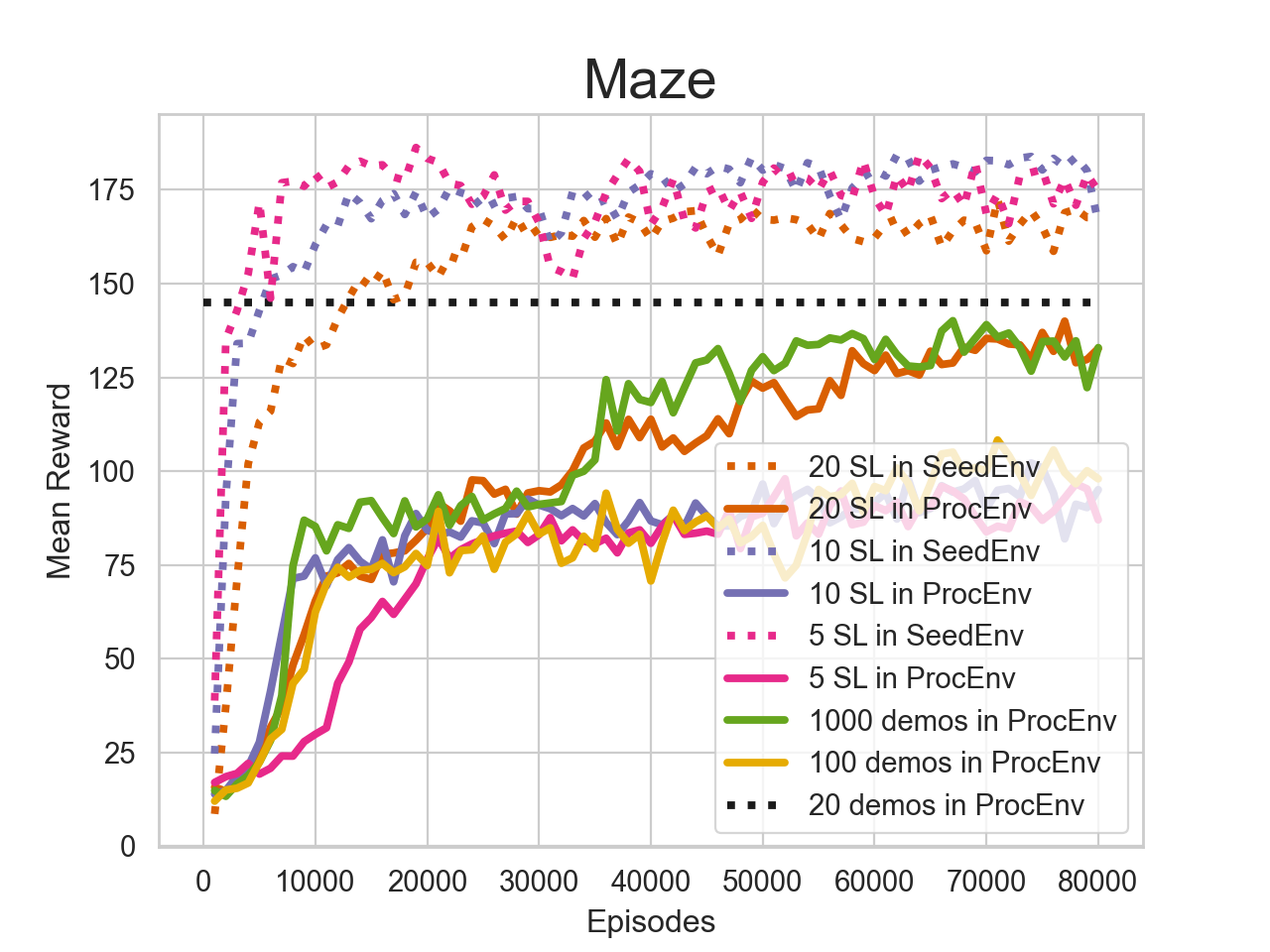} &
		\includegraphics[width=0.37\textwidth]{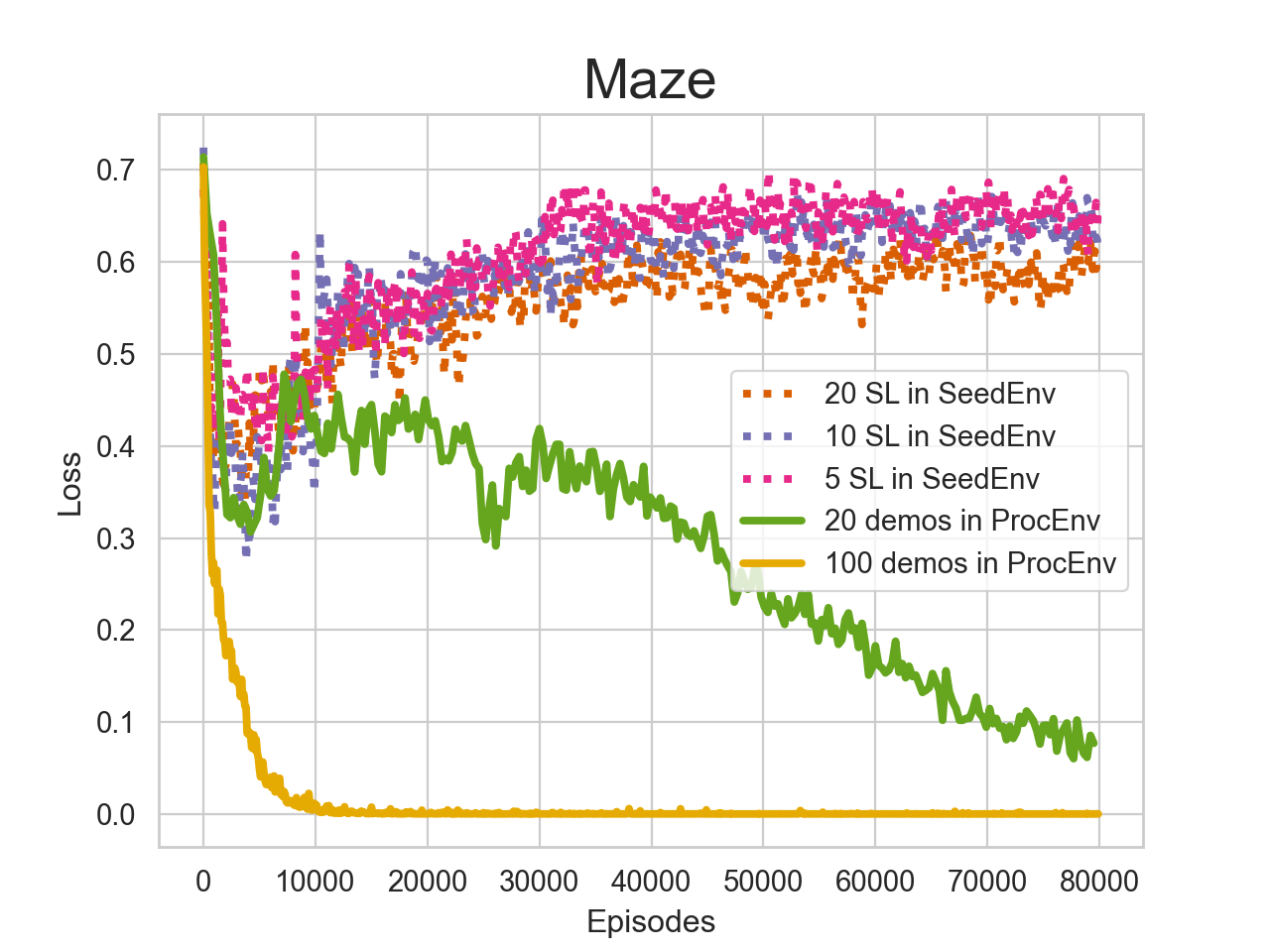} &
		\includegraphics[width=0.37\textwidth]{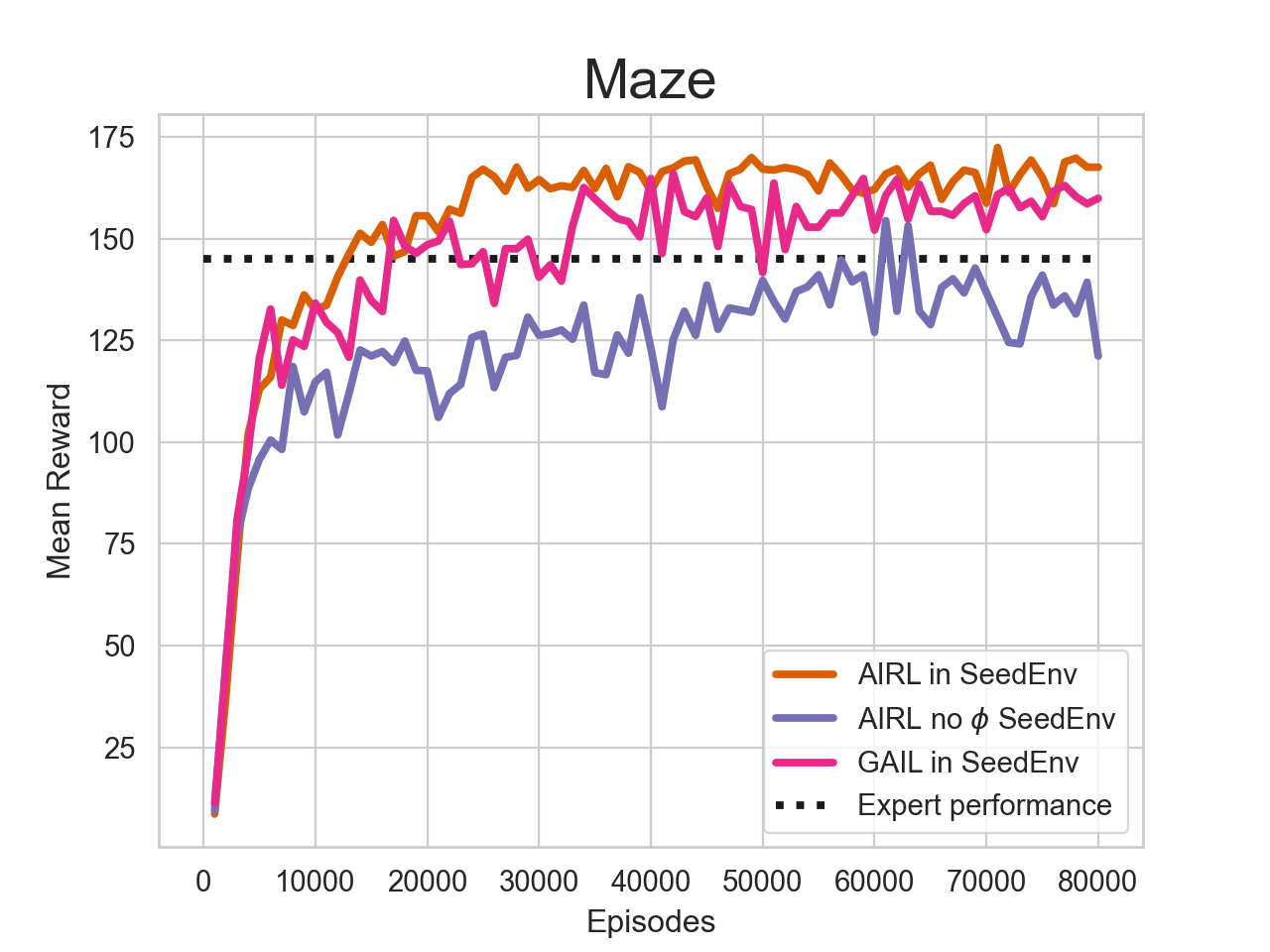} \\

		\includegraphics[width=0.37\textwidth]{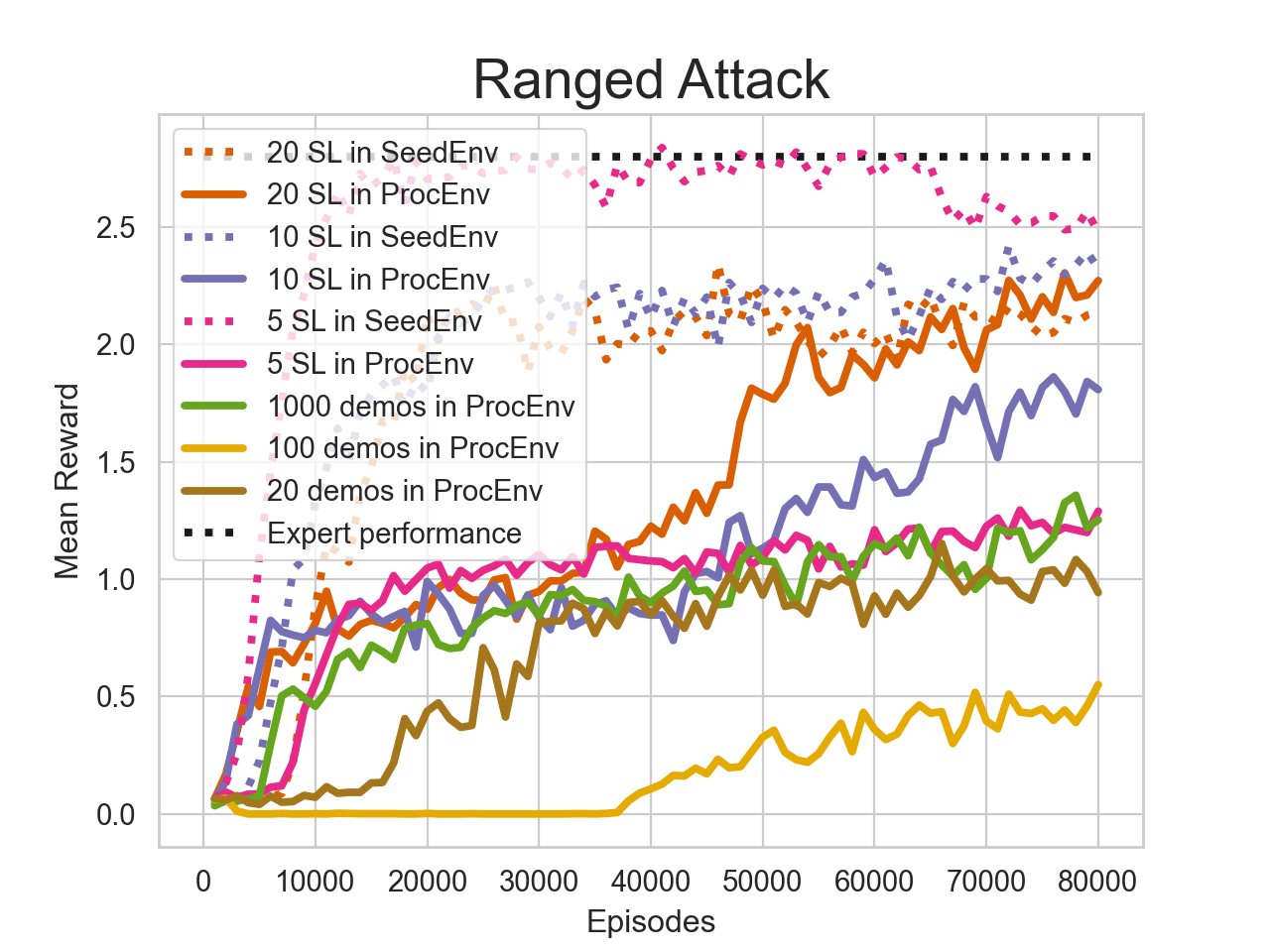} &
		\includegraphics[width=0.37\textwidth]{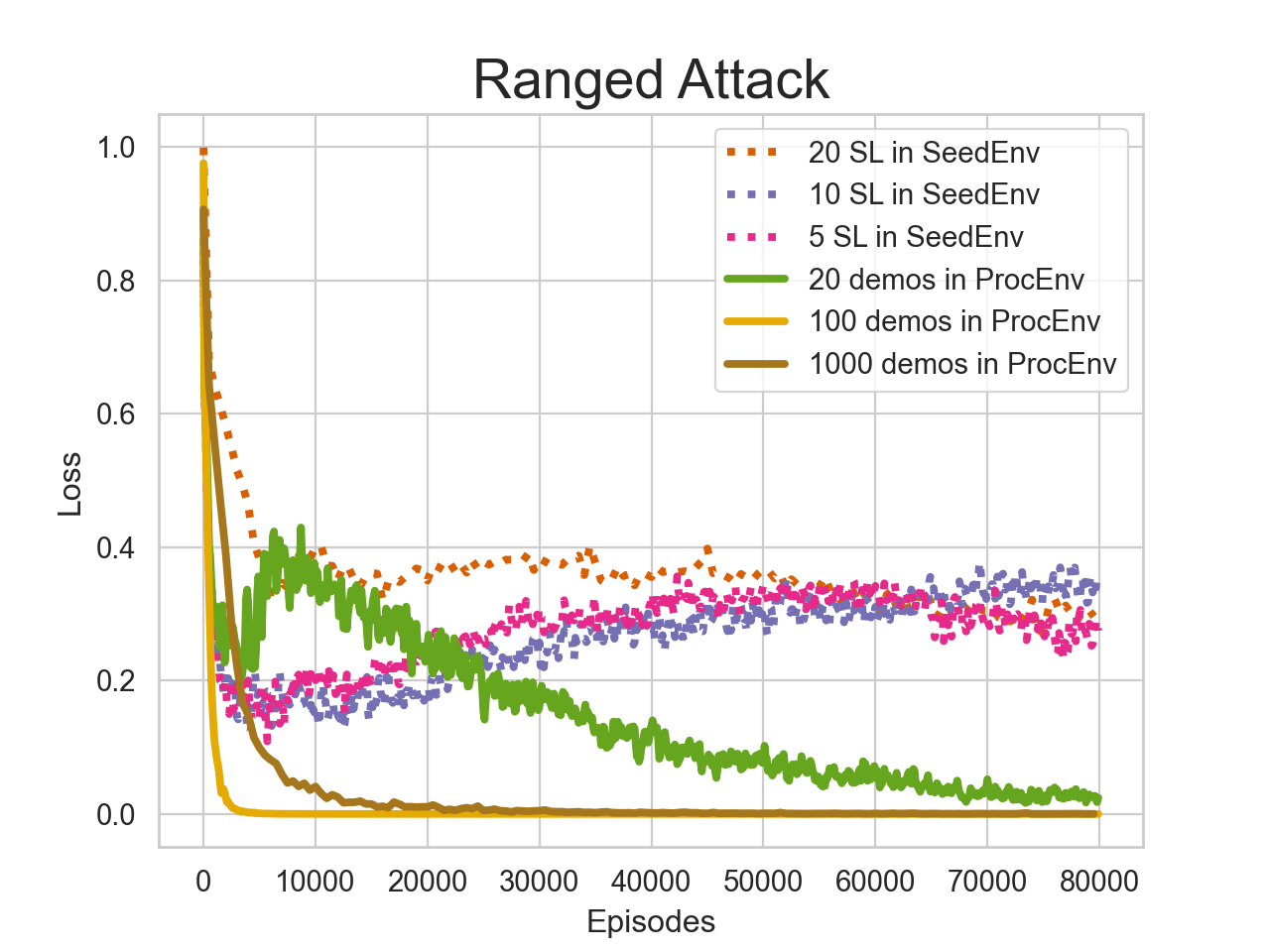} &
		\includegraphics[width=0.37\textwidth]{imgs/mage_all_values.png} \\
		(a) &
		(b) &
		(c) \\

      \end{tabular}
    }
  \end{center}
  \caption{Summary of experimental results. Column (a): reward evolution in
    SeedEnv and ProcEnv with different numbers of seed levels, and naive AIRL on
    ProcEnv. Column (b): the evolution of the loss function. Column (c): the
    training of AIRL without the shaping term and GAIL, both in the SeedEnv with
    20 seed levels for DeepCrawl and 40 seed levels for Minigrid. The dotted
    horizontal line refers to expert performance in the ProcEnv.}
  \label{fig:results}
\end{figure*}

In figure \ref{fig:results} we summarize all the experimental results
described in section 6 of the main text. Included are the performance
of our demonstration-efficient AIRL for all tasks, the evolution of
discriminator losses, and plots showing the importance of using a
disentangling reward function.

\renewcommand{\arraystretch}{1.0}
\renewcommand{\arraystretch}{1.0}
	
\end{document}